\documentclass[conference]{IEEEtran}
\usepackage{times}

\usepackage[numbers] {natbib}
\usepackage{multicol}
\usepackage{multirow}
\usepackage[bookmarks=true]{hyperref}
\usepackage{amsmath,amsfonts}
\usepackage{algorithmic}
\usepackage{array}
\usepackage[caption=false,font=normalsize,labelfont=sf,textfont=sf]{subfig}
\usepackage{textcomp}
\usepackage{stfloats}
\usepackage{url}
\usepackage{verbatim}
\usepackage{graphicx}
\usepackage{float}
\usepackage{kotex}
\usepackage[table,xcdraw]{xcolor}
\usepackage[symbol]{footmisc}
\usepackage{amssymb}
\usepackage{subfloat}    
\usepackage{caption}     
\usepackage{placeins}
\usepackage{afterpage} 
\usepackage{colortbl}
\usepackage{subcaption}
\usepackage{hyperref}
\IEEEoverridecommandlockouts 

\begin{document}

\title{%
\textnormal{\small Published in the Proceedings of Robotics: Science and Systems (RSS) 2025}\\[0.2ex]
RAPID: Robust and Agile Planner \\
Using Inverse Reinforcement Learning \\
for Vision-Based Drone Navigation
}


\author{
Minwoo Kim$^{*, \dagger, 1}$, Geunsik Bae$^{*, 1}$, Jinwoo Lee$^{1}$, Woojae Shin$^{1}$, Changseung Kim$^{1}$, \\ Myong-Yol Choi$^{1}$, Heejung Shin$^{1}$, and Hyondong Oh$^{\dagger\dagger, 1}$%
\thanks{$^{*}$Equal contributions. $^{\dagger}$Project lead. $^{\dagger\dagger}$Corresponding author.}%
\thanks{$^{1}$Department of Mechanical Engineering, Ulsan National Institute of Science and Technology, Republic of Korea.}%
\thanks{Emails: \{red9395, baegs94, jinwoolee2021, oj7987, pon02124, mychoi, godhj, h.oh\}@unist.ac.kr}
}

\maketitle

\begin{abstract}
This paper introduces a learning-based visual planner for agile drone flight in cluttered environments. The proposed planner generates collision-free waypoints in milliseconds, enabling drones to perform agile maneuvers in complex environments without building separate perception, mapping, and planning modules. Learning-based methods, such as behavior cloning (BC) and reinforcement learning (RL), demonstrate promising performance in visual navigation but still face inherent limitations. BC is susceptible to compounding errors due to limited expert imitation, while RL struggles with reward function design and sample inefficiency. To address these limitations, this paper proposes an inverse reinforcement learning (IRL)-based framework for high-speed visual navigation. By leveraging IRL, it is possible to reduce the number of interactions with simulation environments and improve capability to deal with high-dimensional spaces (i.e., visual information) while preserving the robustness of RL policies. A motion primitive-based path planning algorithm collects an expert dataset with privileged map data from diverse environments (e.g., narrow gaps, cubes, spheres, trees), ensuring comprehensive scenario coverage. By leveraging both the acquired expert and learner dataset gathered from the agent's interactions with the simulation environments, a robust reward function and policy are learned across diverse states. While the proposed method is trained in a simulation environment only, it can be directly applied to real-world scenarios without additional training or tuning. The performance of the proposed method is validated in both simulation and real-world environments, including forests and various structures. The trained policy achieves an average speed of 7~m/s and a maximum speed of 8.8~m/s in real flight experiments. To the best of our knowledge, this is the first work to successfully apply an IRL framework for high-speed visual navigation of drones. The experimental videos can be found at \url{https://youtu.be/ZfV6ij0qZMI}.
\end{abstract}

\IEEEpeerreviewmaketitle

\section{Introduction}
Small unmanned aerial vehicles (UAVs), also known as drones, are agile and compact, making them ideal for diverse applications such as search and rescue operations in disaster areas, urban indoor environment exploration, and target tracking. However, utilizing this agility in complex environments (e.g., forests and factories) is still limited due to challenges in perception, control, and real-time motion planning. Thus, to fully exploit agility, the development of agile visual navigation algorithms in complex and unknown environments becomes a necessity.

Classical visual navigation approaches rely on modular architectures that utilize separate perception, mapping, and planning~\citep{zhu2021deep}. These methods are widely adopted due to their interpretability and ease of integration with other modules. However, they incur high computational costs and latency, making them unsuitable for agile drone flight. In contrast, end-to-end neural network-based learning integrates perception, mapping, and planning into a single process, reducing latency and enabling rapid real-time planning~\citep{loquercio2021learning}.

Imitation learning (IL) is a supervised approach whose simplest form, behavior cloning (BC), is popular for its ease of implementation but demands large datasets. With limited data, BC suffers from compounding errors as small mistakes amplify over time~\citep{ross2010efficient} and distribution shift, failing to generalize to unseen states~\citep{garg2021iq}. Dataset Aggregation (DAgger) alleviates these issues by iteratively collecting expert labels in new scenarios to produce a more robust policy, but it hinges on the expert’s ability to label states rapidly; if the expert is slow or computationally heavy, data collection and, thus, training can be significantly delayed~\citep{ross2011reduction,ramrakhya2023pirlnav}.

Unlike BC methods, reinforcement learning (RL) enhances robustness by letting agents interact with the environment and optimize policies via reward maximization. Many studies have successfully applied RL to vision-based flight~\citep{kim2022towards, yu2024mavrl, li2024visfly, bhattacharya2024vision}, but it still suffers from difficult reward design, low sample efficiency, and extensive exploration. Parallel simulators have eased data collection, yet pure vision‑based RL typically requires techniques like privileged learning~\citep{pinto2018asymmetric} or curriculum learning~\citep{wang2021survey} to converge. These issues become even more severe in high‑speed scenarios.

Inverse reinforcement learning (IRL) aims to learn an underlying reward from expert behaviors and then derive an optimal policy from that learned reward. While it shares similarities with the BC method in using expert datasets, IRL can achieve better policies with fewer demonstrations by mitigating compounding errors through sampling new states unseen in the expert dataset. One of the representative IRL methods is generative adversarial imitation learning (GAIL)~\citep{ho2016generative}, which integrates IRL and RL training, making the process faster and more stable. Nevertheless, GAIL struggles with challenges such as mode collapse~\citep{thanh2020catastrophic}, which commonly occurs in adversarial networks, and the biased reward problem~\citep{kostrikov2019discriminator} due to the mistreatment of absorbing states during training.

Recent non-adversarial IRL approaches, such as inverse soft Q-imitation learning (IQ-learning)~\cite{garg2021iq} and least-squares inverse Q-learning (LS-IQ)~\cite{al2023ls}, have made notable progress in mitigating instability and bias issues. For instance, LS-IQ introduces Q-function clipping and handles absorbing states to improve robustness. However, these methods still struggle with the curse of dimensionality when dealing with vision-based continuous action spaces, as they need to consider real-time feasibility checks and precise flight attitude. In such situations, properly defining and managing absorbing states (i.e., goal or collision states) becomes especially difficult, underscoring the need for a specialized approach for high-speed visual navigation tasks.

Beyond algorithmic complexity, vision-based IRL presents additional challenges arising from the need to learn meaningful features directly from raw visual data. Unlike state-based IRL, where low-dimensional and interpretable inputs (e.g., positions and velocities) are used, vision-based IRL methods must handle both reward inference and policy training which are heavily reliant on high-quality feature representations. In learning-based visual navigation tasks, a neural network needs to extract task-relevant information from unstructured and noisy images. This process demands substantial data and careful network design to ensure stable and efficient training. As a result, developing a framework that can effectively learn a robust feature extractor (e.g., autoencoder) is crucial for overcoming these challenges and advancing vision-based IRL.

Even if all these challenges are addressed, applying the trained neural network to real-world scenarios still faces the persistent issue of the sim-to-real gap, a fundamental challenge in learning-based approaches~\cite{loquercio2021learning}. This gap is further exacerbated in vision-based flight, where visual information plays a critical role. In particular, the noise characteristics of visual information vary significantly between the two environments and must be carefully considered. Moreover, discrepancies in drone dynamics between these environments further intensify the sim-to-real gap. Therefore, to achieve robust performance in real environments, these factors must be addressed during training.

\subsection{Contributions} \label{contributions}
In this paper, we propose a learning-based planner called \textit{\textbf{R}obust and \textbf{A}gile \textbf{P}lanner using \textbf{I}nverse reinforcement learning for Vision-Based \textbf{D}rone Navigation (\textbf{RAPID})}, which generates agile collision-free trajectories in cluttered environments. The objective of this paper is to develop a generalizable and sample-efficient visual navigation algorithm using high-dimensional visual information, performing reliably in real-world scenarios without real-world data. The main contributions of this work can be represented as:
\begin{itemize}
\item Development of an inverse soft Q-learning-based framework for high-speed visual navigation that achieves robust and sample-efficient learning without manual reward function design by integrating tailored absorbing state treatment for high-speed scenarios;
\item Introduction of an auxiliary autoencoder loss function to mitigate state complexity from high-dimensional visual inputs, thereby enhancing learning efficiency; and
\item Reduction of the sim-to-real gap by accounting for controller tracking error during training, which yields feasible trajectories accurately tracked on real-world hardware and validates high-speed flight experiments in natural and urban environments at an average speed of 7 m/s.
\end{itemize}

The rest of the paper is organized as follows. Section~\ref{Related Works} introduces conventional and learning-based approaches in vision-based flight. Section~\ref{RAPID} introduces the proposed method. Section~\ref{Simulations} shows simulation environments, dataset acquisition process, and in-depth comparisons with the baseline methods. Section~\ref{Experiments} introduces hardware details, system overview, and experiment results in various real-world scenarios. Section~\ref{Limitation} discusses the limitations of the proposed approach and suggests potential improvements. Finally, Section~\ref{Conclusions} concludes the paper by summarizing the findings and outlining future research directions.

\section{Related Works} \label{Related Works}
\subsection{Classical Methods}
Classical vision-based navigation systems typically employ a sequential pipeline that partitions perception, mapping, planning, and control into separate modules~\cite{tordesillas2021faster, zhou2021raptor}. The workflow begins by converting depth images from onboard cameras into 3D point clouds, which are then aggregated to form volumetric representations such as occupancy grid maps or Euclidean signed distance fields (ESDFs)~\cite{zhou2021raptor}. Next, collision-free trajectories are generated using trajectory-optimization methods, and finally, these trajectories are executed via closed-loop control~\cite{lee2010geometric, Falanga2018}. 

While this modular architecture is straightforward and interpretable, it introduces several significant drawbacks. Discretization artifacts arise due to the finite resolution of grid-based maps, leading to reduced map fidelity. These issues are further exacerbated during high-speed maneuvers, where increased pose-estimation errors can degrade accuracy. Furthermore, the sequential nature of the pipeline imposes cumulative latency, limiting its responsiveness in dynamic and time-critical scenarios. These challenges highlight the need for alternative approaches to improve navigation performance under such conditions.

\subsection{Imitation Learning}
Learning-based methods have emerged as a promising alternative to address the limitations of classical vision-based navigation systems. Unlike module-based methods, learning-based methods focus on generating trajectories or control commands directly from raw image inputs without explicit perception, mapping, and planning modules~\cite{loquercio2021learning, gandhi2017learning, loquercio2018dronet, nguyen2024uncertainty}. One of the most widely-used imitation learning approaches is behavior cloning (BC). BC is popular due to its straightforward implementation and high sample efficiency. However, BC training requires high quality datasets. Studies such as~\cite{gandhi2017learning, loquercio2018dronet} collected datasets in real-world environments, while others, including~\cite{loquercio2021learning, nguyen2024uncertainty}, utilized synthesized data from simulation environments for training.

While BC policies can perform well when high-quality datasets are available, they often suffer from compounding errors and distributional shifts due to overfitting to specific scenarios. To address this,~\cite{loquercio2021learning} applied the DAgger method, which collects additional expert data in unseen states during training. However, this method incurs high costs and is challenging to implement in real-time scenarios where an oracle expert is unavailable.

Another approach extends imitation learning by leveraging privileged information during training to directly optimize the cost associated with generated trajectories, thus training a path generation model~\cite{lu2023lpnet, lu2024you, yang2023iplanner}. For instance, studies such as~\cite{lu2023lpnet, lu2024you} calculate Q-functions based on map data to update policies without explicit labeling. On the other hand,~\cite{yang2023iplanner} employs a differentiable cost map to optimize trajectory quality directly, without relying on Q-function learning or reinforcement signals. This method focuses on efficient optimization of path generation under given constraints and can be effective as it operates without explicit labeled data. However, it still faces challenges such as reliance on the quality of the cost map and computational overhead, which may limit its scalability.

\subsection{Reinforcement Learning}
Reinforcement learning (RL) has demonstrated remarkable results across various domains and has shown promise even in challenging fields such as drone visual navigation. Recent studies have explored end-to-end learning approaches that utilize visual data to directly generate low-level control commands~\cite{yu2024mavrl, bhattacharya2024vision, xing2024bootstrapping,  song2023learning, kulkarni2024reinforcement}.

However, RL methods that rely on raw visual information often suffer from slow convergence and require a large amount of data to train. Moreover, the design of effective reward functions poses a significant challenge, as it requires careful consideration to ensure alignment with the desired behaviors and to avoid unintended consequences. These limitations necessitate powerful parallel simulation environments capable of providing diverse state information to train robust policies for various environments~\cite{makoviychuk2isaac, song2021flightmare, kulkarni2023aerial}. Despite these advancements, training vision-based RL policies remains a challenging task, prompting researchers to propose alternative methods to address these difficulties.

For instance, Xing et al.~\cite{xing2024bootstrapping} employed a DAgger-based policy as a foundation and refined it using RL for effective state embedding. Song et al.~\cite{song2023learning} introduced a framework where a state-based RL policy is first trained, followed by knowledge distillation to transfer the knowledge into a vision-based RL policy. Similarly, Bhattacharya et al.~\cite{bhattacharya2024vision} developed a neural network combining vision transformers (ViT) with LSTM to achieve efficient state embeddings.

In drone racing, policies based on raw pixel data have also been investigated~\cite{song2023reaching, geles2024demonstrating}. While these methods demonstrate promising results in constrained racing track environments, their applicability to more diverse and unstructured scenarios, such as natural or urban environments, has not been fully established. Consequently, developing RL methods that effectively utilize raw visual inputs while ensuring robust generalization and fast convergence remains an open and significant research challenge.

\subsection{Inverse Reinforcement Learning}
Inverse reinforcement learning (IRL) aims to find a proper reward from expert samples. IRL is particularly beneficial in applications where reward design is difficult, yet its adaptation to vision-based tasks remains a significant challenge~\cite{ho2016generative}. While some studies have successfully applied IRL to autonomous driving~\cite{lee2021approximate, liu2022improved}, its application to drones is still unexplored. Compared with autonomous driving, autonomous drone navigation is more demanding as it involves 3D spatial awareness and attitude control-including pitch, roll, and yaw-thus making policy learning considerably more complex. Furthermore, drone navigation necessitates highly accurate action generation as the drone is at significant risk of crashing if the action is not generated with sufficient precision. This challenge becomes even greater in high-speed drone flight~\cite{arora2021survey}, where raw sensory visual data exacerbate the difficulty of reward shaping and policy learning. Moreover, drones are highly sensitive to external factors such as wind disturbances, sensor noise, and limited onboard computational resources, adding further complexity to the effective use of IRL. Consequently, direct application of IRL to vision-based high-speed drone flight remains a significant challenge.

\section{Methodology} \label{RAPID}
\begin{figure*}[!ht]
    \centering
    \includegraphics[width=\textwidth]{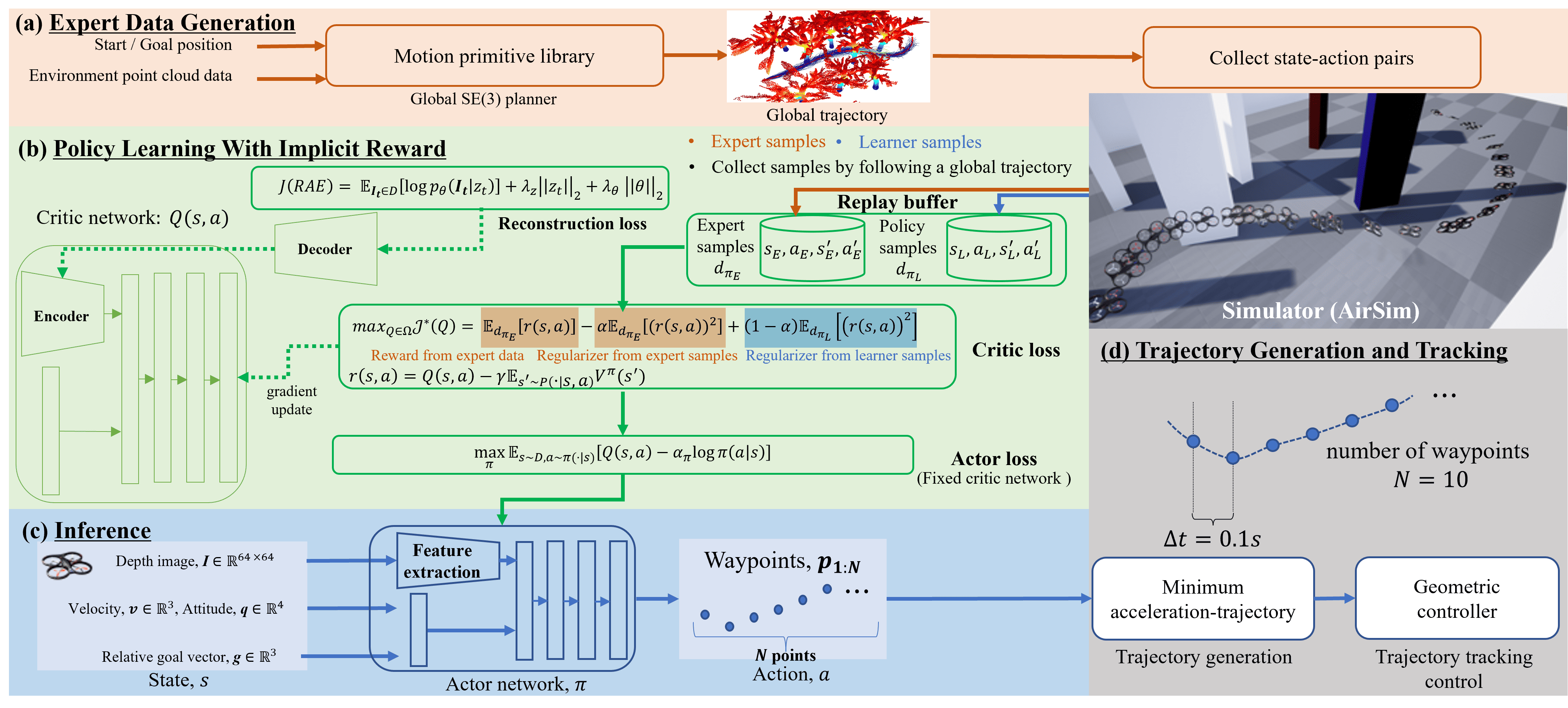}
    \caption{Overview of the learning framework for the proposed inverse soft Q-imitation learning method. The learning framework is composed of four parts: (a) expert data generation, (b) policy learning with implicit reward, (c) inference, and (d) trajectory generation and tracking.}
    \label{fig:training_framework}
\end{figure*}

RAPID is a vision-based planner utilizing IRL, designed to generate waypoints from depth images and drone state inputs. These waypoints are converted into continuous trajectories and executed by a tracking controller. The training process, from dataset acquisition to action implementation in simulation, is illustrated in Fig.~\ref{fig:training_framework}. Section~\ref{States and Actions} outlines the states and actions for RAPID training, while Section~\ref{Sample Efficient} presents an auxiliary loss function and network structure for sample-efficient learning. Section~\ref{Policy Learning With Implicit Reward} details reward inference and policy updates in the IRL framework. Finally, Section~\ref{Trajectory Generation and Tracking Control} explains trajectory generation and tracking control.

\subsection{Preliminaries}
The vision-based navigation problem can be modeled as an infinite-horizon Markov decision process (MDP). The MDP is composed of~$(s, a, p(s_{0}), s', p(s', s|a), r(s, a), \gamma)$, where $s$ is a state, $a$ is an action, $p(s_{0})$ is an initial state distribution, $s'$ is a next state, $p(s', s|a)$ is a transition probability, $r(s, a)$ is a reward and $\gamma \in [0, 1]$ is a discount factor. $\pi$ is a stochastic policy that takes an action $a$ given a state $s$. Data from the expert policy will be denoted as $D_{\pi_{E}}$, while data from the learner policy will be denoted as $D_{\pi}$. Additionally, the expert data distribution is represented as $d_{\pi_{E}}$, and the learner data distribution is represented as $d_{\pi}$. A detailed explanation of states and actions is provided in the following section.

\subsection{States and Actions} \label{States and Actions}
\subsubsection{States} \label{States}
The policy network $\pi(a_t|s_t)$ generates an action $a_t$ at time step $t$. The state space $s_t$ is defined as:
\begin{equation*}
    \boldsymbol{s_t} = [\boldsymbol{\mathit{I_t, v_t, q_t, g_t}}],
\end{equation*}
where a depth image \textbf{\textit{I}} $ \in \mathbb{R}^{64 \times 64}$, velocity \textbf{\textit{v}} $ \in \mathbb{R}^{3}$, attitude quaternion \textbf{\textit{q}} $ \in \mathbb{R}^{4}$, and a relative goal vector \textbf{\textit{g}} $ \in \mathbb{R}^{3}$ (i.e., the difference between the goal and the current position of the drone). 

\begin{figure}
    \centering
    \includegraphics[width=\columnwidth]{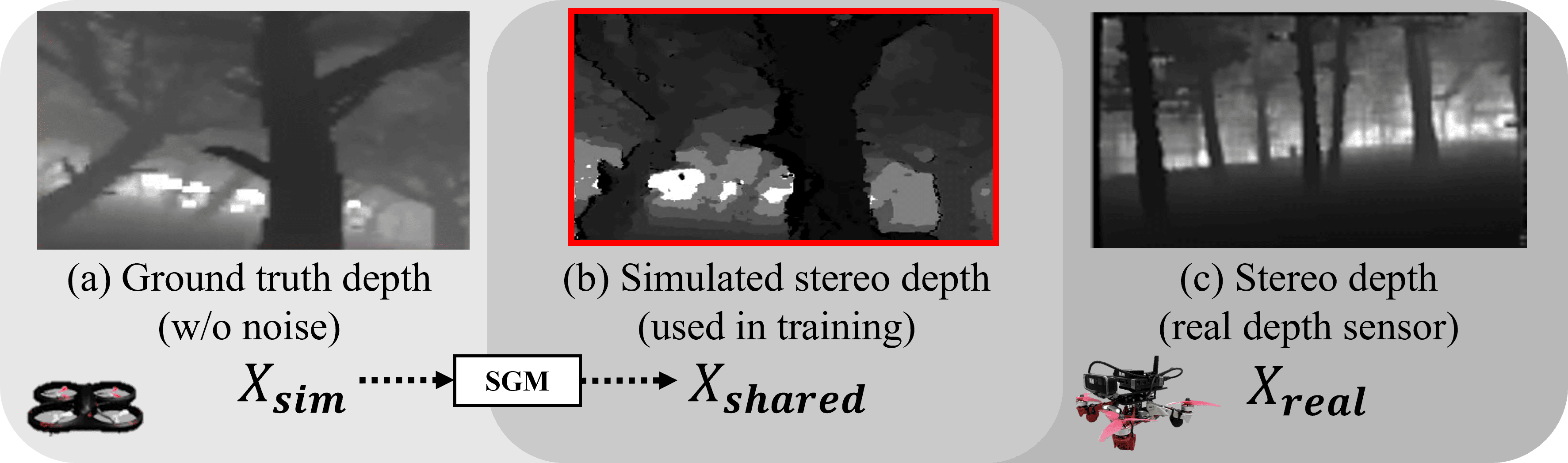}
    \caption{Depth images from simulation and real-world: (a) ground truth depth image, (b) simulated stereo depth image, and (c) stereo depth image from a real depth sensor. To reflect realistic sensor noise during training, a simulated stereo depth image is generated through a stereo vision algorithm.}
    \label{fig:real_experiment_gray_depth}    
\end{figure}

In general, domain-invariant features such as depth images are used to bridge the gap between simulation and real-world environments. However, as shown in Fig.~\ref{fig:real_experiment_gray_depth}, there are differences between the depth images from the simulation and real-world. Therefore, these differences need to be addressed to overcome the sim-to-real gap. To this end, a stereo depth image similar to a real depth image is calculated through the semi-global matching (SGM~\cite{hirschmuller2008stereo}) method and used for training (Fig.~\ref{fig:real_experiment_gray_depth}(b)). 

While high-resolution images generally improve performance in simulation environments, they require larger network architectures and can lead to overfitting, resulting in reduced success rates during testing~\cite{zhang2024back}. This overfitting may increase dependency on high-quality depth information during inference, which is impractical for real-world scenarios where depth maps from a sensor are often noisy and inconsistent. To address this, lower-resolution images ($64 \times 64$) are used to reduce overfitting and improve robustness, thereby narrowing the gap between simulation and real-world environments.

\subsubsection{Actions} \label{Actions}

\begin{figure}[!t]
    \centering
    \includegraphics[width=\columnwidth]{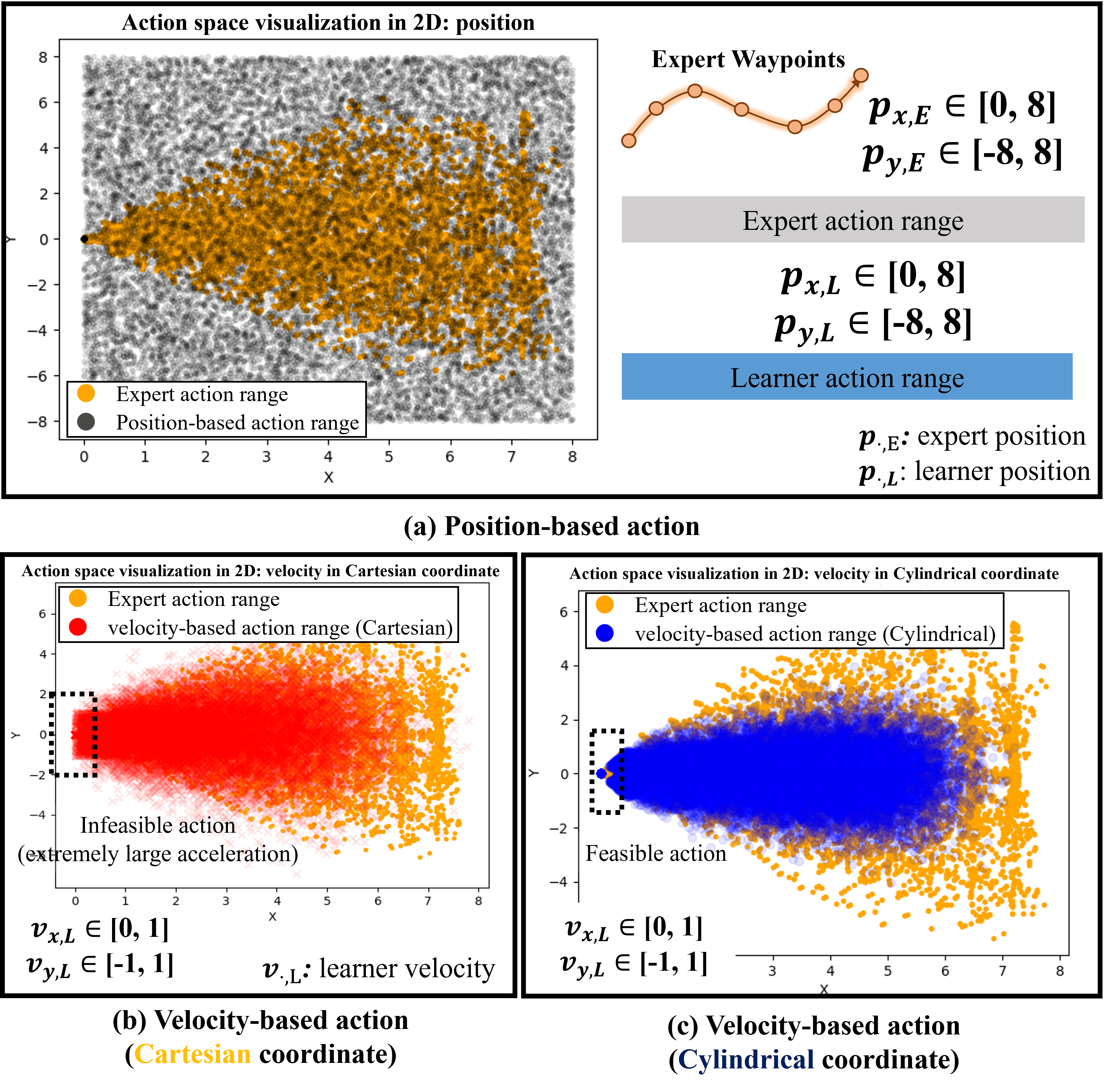}
    \caption{Action space representation with different coordinate systems: (a) position-based action space, (b) velocity-based action space in Cartesian coordinates, and (c) velocity-based action space in cylindrical coordinates. The action space generated by the neural network is referred to as the \textbf{learner} action to distinguish it from the expert action.}
    \label{fig:action space}
\end{figure}

The action $a_t$ consists of $N$ waypoints ahead, each separated by a fixed time interval $T$. Each waypoint represents a relative position from the previous one, expressed in cylindrical coordinates to reduce the complexity of the action space. For clarity, the action generated by the policy network is referred to as the raw action $a^{\text{raw}}$, while the post-processed action is denoted as $a$. Specifically, each waypoint in the raw action $a^{\text{raw}}$ is defined by a relative distance $\Delta r_i$ and a relative angle $\Delta \psi_i$. The raw action $a^{\text{raw}}$ generated from the policy network is:
\[
a_t^{raw} = \left\{ (\Delta r_1, \Delta \psi_1), (\Delta r_2, \Delta \psi_2), \dots, (\Delta r_N, \Delta \psi_N) \right\}.
\]
These waypoints are transformed into Cartesian coordinates using cumulative heading angles. Let \( \theta_0 = \psi_t \) be the initial heading angle of the drone. The cumulative heading angle for the \( i \)-th waypoint is defined as:
\[
\theta_i = \theta_{i-1} + \Delta \psi_i, \quad \text{for } i = 1, 2, \dots, N.
\]
The position of the \( i \)-th waypoint in Cartesian coordinates is then calculated recursively:
\[
\mathbf{p}_i = \mathbf{p}_{i-1} + \Delta r_i
\begin{bmatrix}
\cos \theta_i \\
\sin \theta_i
\end{bmatrix},
\quad \text{with } \mathbf{p}_0 = \begin{bmatrix} x_t \\ y_t \end{bmatrix}.
\]
Accordingly, the final transformed action \( a_t \) is:
\[
a_t = \left\{ \mathbf{p}_1, \mathbf{p}_2, \dots, \mathbf{p}_N \right\},
\]
where each \( \mathbf{p}_i \) is the absolute position of the \( i \)-th waypoint in Cartesian coordinates. In this paper, $N$ is set to 10 and a fixed time interval of $T=0.1s$ is used.

Figure~\ref{fig:action space} illustrates the action space with different coordinate systems. As shown in Fig.~\ref{fig:action space}(a), position-based actions in Cartesian coordinates can encompass the expert action range but significantly increase the search space, making training highly challenging. To address this, Fig.~\ref{fig:action space}(b) demonstrates velocity-based actions in Cartesian coordinates, which partially reduce the search space but still result in infeasible trajectories at the beginning (e.g., trajectories with excessive acceleration). In contrast, cylindrical coordinates, as shown in Fig.~\ref{fig:action space}(c), align the action range more closely with that of the expert, eliminating initial infeasible actions. While the proposed action representation may appear unconventional, it effectively confines the action range within a specific boundary and stabilizes training~\cite{kanervisto2020action}.

\subsection{Sample Efficient Training With Image Reconstruction} \label{Sample Efficient}
\subsubsection{Auxiliary Loss Function for Autoencoder}
Vision-based RL faces significant challenges due to the complexity of processing high-dimensional visual inputs. Unlike state-based RL, vision-based RL requires efficient representation learning to extract meaningful features, often resulting in lower sample efficiency and longer training time. The stochastic nature of visual data and the risk of overfitting further complicate learning. These limitations make generalization difficult and necessitate auxiliary tasks or separate robust feature extraction methods to improve performance. To address these challenges, a $\beta$-variational autoencoder ($\beta$-VAE~\cite{kingma2014auto}) is utilized to learn compact state representations, effectively embedding high-dimensional inputs while mitigating noise and improving robustness in visual data processing. $\beta$-VAE consists of two components: a convolutional encoder $g_{\phi}$, which maps an image observation $I_t$ to a low-dimensional latent vector $z_t$, and a deconvolutional decoder $f_{\theta}$, which reconstructs $z_t$ back to the original state $I_t$. To stabilize training and enhance performance, an $\ell_2$ penalty is applied to the learned representation $z_t$, and weight decay is imposed on the decoder parameter $\theta$ as auxiliary objectives. The objective function of the reconstruction autoencoder (RAE), denoted as \( J(RAE) \), is given as:
\begin{equation}
        J(RAE) = \mathbb{E}_{I_t \sim D}[\log p_{\theta}(I_t|z_t) + \lambda_z||z_t||_{2} + \lambda_{\theta}||\theta||_{2}],
\end{equation}
where $z_t = g_{\phi}(I_t)$ and $\lambda_z$ and $\lambda_{\theta}$ are hyperparameters. 

\begin{figure}[t]
    \centering
    \includegraphics[width=\columnwidth]{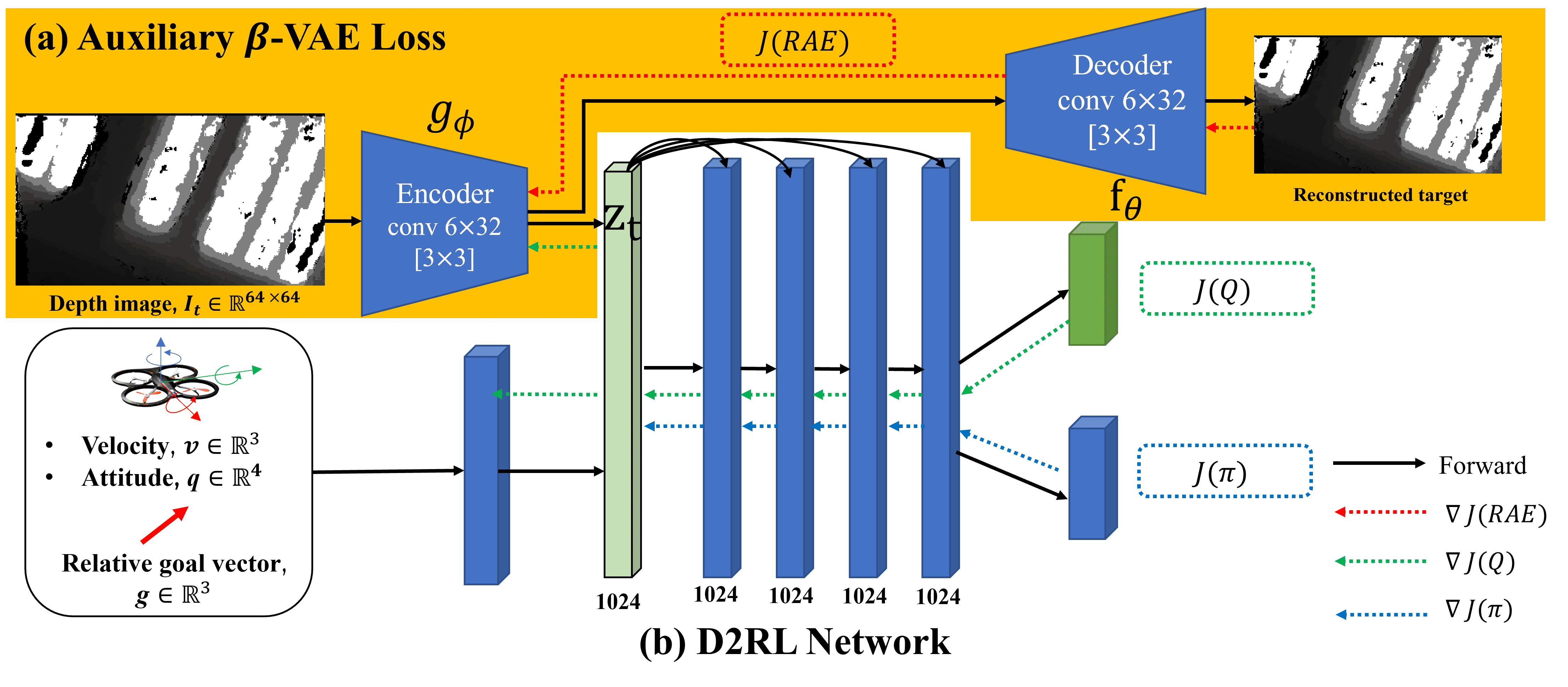}
    \caption{Auxiliary reinforcement learning using autoencoder and skipping connection networks.}
    \label{fig:auxiliary loss}
\end{figure}

Following the approach proposed in prior work~\cite{yarats2021improving}, we adopt a strategy where the actor and critic share the same convolutional encoder parameters to process high-dimensional inputs. However, prior work has also shown that allowing the actor's gradients to update the shared encoder can negatively impact the agent's performance. This issue arises because the encoder is shared between the policy $\pi$ and the Q-function, causing updates from the actor's gradients to unintentionally alter the Q-function's representation. To address this, we block the actor's gradients from propagating to the encoder. In other words, the encoder is updated solely using the critic's gradients, as the Q-function contains all task-relevant information. This approach stabilizes learning and ensures that the encoder effectively learns task-dependent representations without interference from the actor's updates. The proposed approach is illustrated in Fig.~\ref{fig:auxiliary loss}(a).
 
However, the restriction of encoder update from the actor significantly slows down the encoder’s updates. To further address the delayed signal propagation caused by restricting actor gradients, we apply a faster Polyak averaging rate ($\rho$) to the encoder parameters of the target Q-function compared with the rest of the networks. These strategies ensure robust and efficient learning while maintaining stable task-dependent feature representations. In the original work~\cite{yarats2021improving}, the use of larger Polyak averaging rates ($\rho_{\text{enc}} = 0.05$ and $\rho_Q= 0.01$) yielded reasonable performance in simple tasks; however, in high-speed flight task which requires precise actions, we observed that smaller averaging rates ($\rho_{\text{enc}} = 0.01$ and $\rho_Q = 0.005$) improved performance. 

\subsubsection{Skipping Connection Networks}
Although deeper networks generally perform better on complex tasks by introducing inductive biases, simply adding more layers in RL does not yield the same benefits as in computer vision tasks. This is because additional layers can decrease mutual information between input and output due to non-linear transformations. To overcome this issue, skip connections can be used to preserve important input information and enable faster convergence. We applied the deeper dense RL (D2RL)~\cite{sinha2020d2rl} network, which incorporates such skip connections into RL, to the high-speed visual navigation problem. This allows us to gain the advantages of deeper neural networks while achieving faster learning. The skip connection network is illustrated in Fig.~\ref{fig:auxiliary loss}(b).

Furthermore, it is discovered that network initialization plays an important role during the learning process. Commonly used initialization techniques, such as Xavier initialization~\cite{glorot2010understanding}, often lead to instability in learning. Specifically, we initialize the weight matrix of fully-connected layers using orthogonal initialization~\cite{saxe2013exact} with zero bias, while convolutional and deconvolutional layers are initialized with delta-orthogonal initialization~\cite{xiao2018dynamical}. Detailed network structure and learning hyperparameters can be found in the appendix (Section~\ref{Appendix}).

\subsection{Policy Learning With Implicit Reward} \label{Policy Learning With Implicit Reward}
This section introduces the IRL-based policy update method, covering: 1) reward and Q-function learning using inverse soft Q-learning, 2) managing absorbing states in high-speed visual navigation, and 3) policy updates via soft actor-critic.

\subsubsection{Learning Implicit Reward}
The IRL algorithm utilized in this study is least squares inverse Q-learning (LS-IQ)~\cite{al2023ls}, which directly learns a Q-function through an implicit reward formulation. Previously, IRL methods required simultaneous training of two neural networks in the reward-policy domain. However, inverse soft Q-imitation learning (IQ-learning~\cite{garg2021iq}) introduced the inverse Bellman operator $\mathcal{T}^\pi Q$, enabling a mapping from the reward function domain to the Q-function domain. This innovation allows rewards to be expressed entirely in terms of Q-functions, eliminating the need for explicit reward network training. The inverse Bellman operator $\mathcal{T}^\pi Q$, following a policy $\pi$, is defined as:
\begin{equation} \label{eq:Inverse Bellman operator}
    (\mathcal{T}^\pi Q)(s,a) = Q(s,a) - \gamma \mathbb{E}_{s' \sim P(\cdot|s,a)} V^\pi(s'),
\end{equation}
where $V^\pi(s)$ is the value function following policy $\pi$, defined as $V^\pi(s) = \mathbb{E}_{a \sim \pi(\cdot|s)}[Q(s,a) - \log \pi(a|s)]$. From Eq.~(\ref{eq:Inverse Bellman operator}), the reward function can be expressed as $r(s,a) = \mathcal{T}^\pi Q$, allowing simultaneous optimization of the Q-function and rewards.

Building on this framework, the IRL problem is transformed into a single maximization objective, $J(Q,\pi)$. Specifically, the use of the inverse Bellman operator reformulates the optimization problem from the reward-policy space to the Q-policy space:
\begin{equation*}
    \max_{r \in R} \min_{\pi \in \Pi} L(r,\pi) = \max_{Q \in \Omega} \min_{\pi \in \Pi} J(Q,\pi),
\end{equation*}
where $\Omega = \mathbf{R}^{S \times A}$ denotes the space of Q-functions. From the soft-Q learning concept~\cite{haarnoja2018soft}, given a Q-function is fixed, the optimal policy can be expressed as $\pi_{Q}(a|s) = \frac{1}{Z_s}\exp Q(s, a),$ where $Z_s$ is a normalization factor defined as $Z_s=\sum_{a}\exp Q(s, a)$. Leveraging this formulation, the optimization objective simplifies to learning only the Q-function:
\begin{equation}\label{IQL objective}
    \max_{Q \in \Omega} \min_{\pi \in \Pi} \mathit{J}(\pi, Q) = \max_{Q \in \Omega} \mathit{J}(\pi_{Q}, Q).
\end{equation}
This transformation accelerates learning by eliminating the need for a separate reward network.

To enhance the learning stability of $J(Q, \pi)$, the algorithm employs a regularizer $\psi(r)$, which imposes constraints on the magnitude and structure of the Q-function. This helps to prevent overfitting, ensures stable learning, and improves generalization. In practice, $\ell_2$ regularization can be applied to enforce a norm penalty, leveraging a $\chi^2$-divergence. IQ-learning~\cite{garg2021iq} applies an $\ell_2$ norm-penalty on the reward function over state-action pairs sampled from expert trajectories. However, this approach has been shown to cause instability in continuous action spaces.

To address this issue, LS-IQ~\cite{al2023ls} stabilizes learning by introducing a mixture of distributions from both expert and learner data. The regularizer $\psi(r)$ is defined as:
\begin{equation*}
    \psi(r) = \alpha \mathbb{E}_{d_{\pi_{E}}}[r(s,a)^2] + (1-\alpha) \mathbb{E}_{d_{\pi_{L}}}[r(s,a)^2],
\end{equation*}
where $\alpha$ is set to $0.5$. This mixture-based regularization mitigates instability by balancing contributions from expert and learner distributions. Consequently, the Q-function objective $J(Q,\pi)$ is expressed as:
\begin{equation} \label{eq:Q-function objective}
\begin{aligned}
    J(Q, \pi) &= \mathbb{E}_{d_{\pi_E}}[r(s,a)] - \alpha \mathbb{E}_{d_{\pi_E}}[(r(s,a))^2]  \\            &- (1-\alpha)\mathbb{E}_{d_{\pi_L}}[(r(s,a))^2] \\
              &- \mathbb{E}_{d_{\pi_L} \cup d_{\pi_E}}[V^\pi(s) - \gamma \mathbb{E}_{s' \sim P(\cdot|s,a)}  V^\pi(s')], 
\end{aligned}
\end{equation}
where $r(s,a) = Q(s,a) - \gamma \mathbb{E}_{s' \sim P(\cdot|s,a)} V^\pi(s')$ as explained in Eq.~(\ref{eq:Inverse Bellman operator}). The last term of Eq.~(\ref{eq:Q-function objective}) removes the state bias.

Furthermore, LS-IQ enhances stability in implicit learning methods by effectively handling absorbing states and applying Q-function clipping during training. The inverse Bellman operator $\mathcal{T}^\pi Q$, accounting for an absorbing state $s_A$, is defined as:
\begin{equation*}
    \mathcal{T}^\pi Q(s,a) = Q(s,a) - \gamma \mathbb{E}_{s' \sim P(\cdot|s,a)}\big((1-\nu)V^\pi(s') + \nu V(s_A)\big),
\end{equation*}
where $\nu$ is an indicator such that $\nu = 1$ if $s'$ is a terminal state and $\nu = 0$ otherwise. The value of the absorbing state \( V(s_A) \) is computed in closed form as \( V(s_A) = \frac{r_A}{1-\gamma} \), representing the total discounted return under an infinite time horizon, where \( r_A \) is set to \( r_{\text{max}} \) for expert states and \( r_{\text{min}} \) for learner states.

In our settings, $r_{\text{max}}$ and $r_{\text{min}}$ are calculated as 2 and $-2$, respectively. The mathematical definitions and proofs of \( r_{\text{max}} \) and \( r_{\text{min}} \) are detailed in the original paper, and readers are referred to~\cite{al2023ls} for further elaboration. The value $V(s_{A})$ is mathematically bounded and can be computed either analytically or via bootstrapping. In this paper, the LS-IQ method adopts the bootstrapping approach for updates. The full objective of Eq.~(\ref{eq:Q-function objective}) including terminal state treatment is shown in the appendix (Section~\ref{Appendix}).

\subsubsection{Managing Absorbing States in High-Speed Visual Navigation}
In problems like high-speed visual navigation, where terminal states (e.g., goal or collision states) frequently appear, recursive bootstrapping for absorbing states often causes instability. To resolve this issue, we propose a method that combines bootstrapping for non-terminal states with analytical computation for absorbing states, resulting in a significant improvement in stability and overall performance.

Along with refining the computation method for state values, we also adjust the values of $r_{\text{max}}$ and $r_{\text{min}}$ to better suit the high-speed visual navigation scenario. During our initial experiments, we set $r_{\text{max}} = 2$. However, this configuration caused instability during training as the agent received high rewards upon reaching terminal states, even when dangerously close to obstacles. To minimize this effect, we asymmetrically set $r_{\text{max}} = 0$ and $r_{\text{min}} = -2$. This adjustment prevented undesired high rewards in terminal states and significantly enhanced obstacle avoidance performance.

\subsubsection{Soft Actor-Critic Update}
To train a policy, soft actor-critic (SAC~\cite{haarnoja2018soft}) is used. In the continuous action space, there is no direct way to get an optimal policy. Instead, an explicit policy $\pi$ is used to approximate $\pi_{Q}$ by using the SAC method. With a fixed Q-function, the policy $\pi$ is updated using the following equation:
\begin{equation} \label{Actor loss}
    \max_{\pi} \mathbb{E}_{s \sim D, a \sim \pi(\cdot|s)}[Q(s,a) - \alpha_{\pi} \log \pi(a|s)],   
\end{equation}
where $D$ is a replay buffer, and $\alpha_{\pi}$ is the temperature parameter. The temperature $\alpha_{\pi}$ controls the trade-off between exploration and exploitation by scaling the entropy term.

\subsection{Trajectory Generation and Control} \label{Trajectory Generation and Tracking Control}
Given discrete waypoints generated by the network, it is necessary to convert them into a continuous and differentiable trajectory for smooth flight. The trajectory $\boldsymbol{\tau}(t)$ can be represented as a distinct function along each axis as:
\begin{align*}
\boldsymbol{\tau}(t):=\left[\tau_x(t), \tau_y(t), \tau_z(t)\right]^T.
\end{align*}
For each axis, the trajectory can be represented as an $M^{th}$-order of piecewise polynomial function with $N$ time intervals:

\begin{gather*}
\tau_{\mu}^{k}(t) = \mbox{$\sum_{i=0}^M$} \sigma_{\mu, i}^{k}\left(t-t_0-kT\right)^i, \\
t_0 + (k-1)T \leq t < t_0 + kT, 
\end{gather*}
where $\mu\in\{x,y,z\}$ and $k = 1,\cdots,N$.

Each polynomial segment must start and end at specified waypoints and ensure a smooth transition by maintaining the continuity of the $j^{th}$ derivative at each intermediate waypoint. Moreover, the first segment should initiate from the drone's current position, velocity, and acceleration. 
The trajectory that minimizes the integral of the acceleration squared can be found by solving the optimization problem as:
\begin{align*}
&\mbox{$ \underset{\boldsymbol{\sigma}^1,\boldsymbol{\sigma}^2,\cdots,\boldsymbol{\sigma}^N}{\mathrm{min}}\; J = \int_{t_0}^{t_N} \|\ddot{\boldsymbol{\tau}}(t)\|^2 dt $},
\end{align*}
where ~\mbox{$\boldsymbol{\sigma}^k~\in ~\mathbb{R}^{(M+1)\times3}$} represents the coefficients of the $k^{th}$~polynomial segment. The objective $J$ can be analytically computed by integrating the polynomial, and it is formulated as constrained quadratic programming (QP). In this study, we adopt the method from~\cite{richter2016polynomial}, which provides a closed-form solution by mapping the polynomial coefficients to the derivatives at each segment boundary. We employ a 4th-degree polynomial (i.e., $M=4$) with velocity continuity at all intermediate waypoints (i.e., $j=1$) and impose zero terminal velocity and acceleration at the final waypoint.

The generated trajectories can be executed via closed-loop control~\cite{lee2010geometric, Falanga2018}, with model predictive control (MPC~\cite{Falanga2018}) or geometric controllers~\cite{lee2010geometric} being two of the most commonly used methods. MPC generates safe and feasible trajectories by solving an optimization problem under predefined constraints, providing strong stability properties but at the expense of high computational complexity. While MPC is fast enough for real-time control, its computational overhead becomes a bottleneck in RL training due to the need for rapid, repeated evaluations. In contrast, geometric controllers ensure tracking accuracy and stability by directly applying geometric principles of rigid-body dynamics, resulting in significantly lower computational overhead. Because of its low latency and ease of implementation, a geometric controller is more suitable for the learning process compared with MPC. Consequently, this work adopts a geometric controller for trajectory tracking.

\section{Simulations} \label{Simulations}
\subsection{Data Acquisition and Training}
\subsubsection{Data Acquisition} 
\begin{figure}[!t]
    \centering
    \includegraphics[width=\columnwidth]{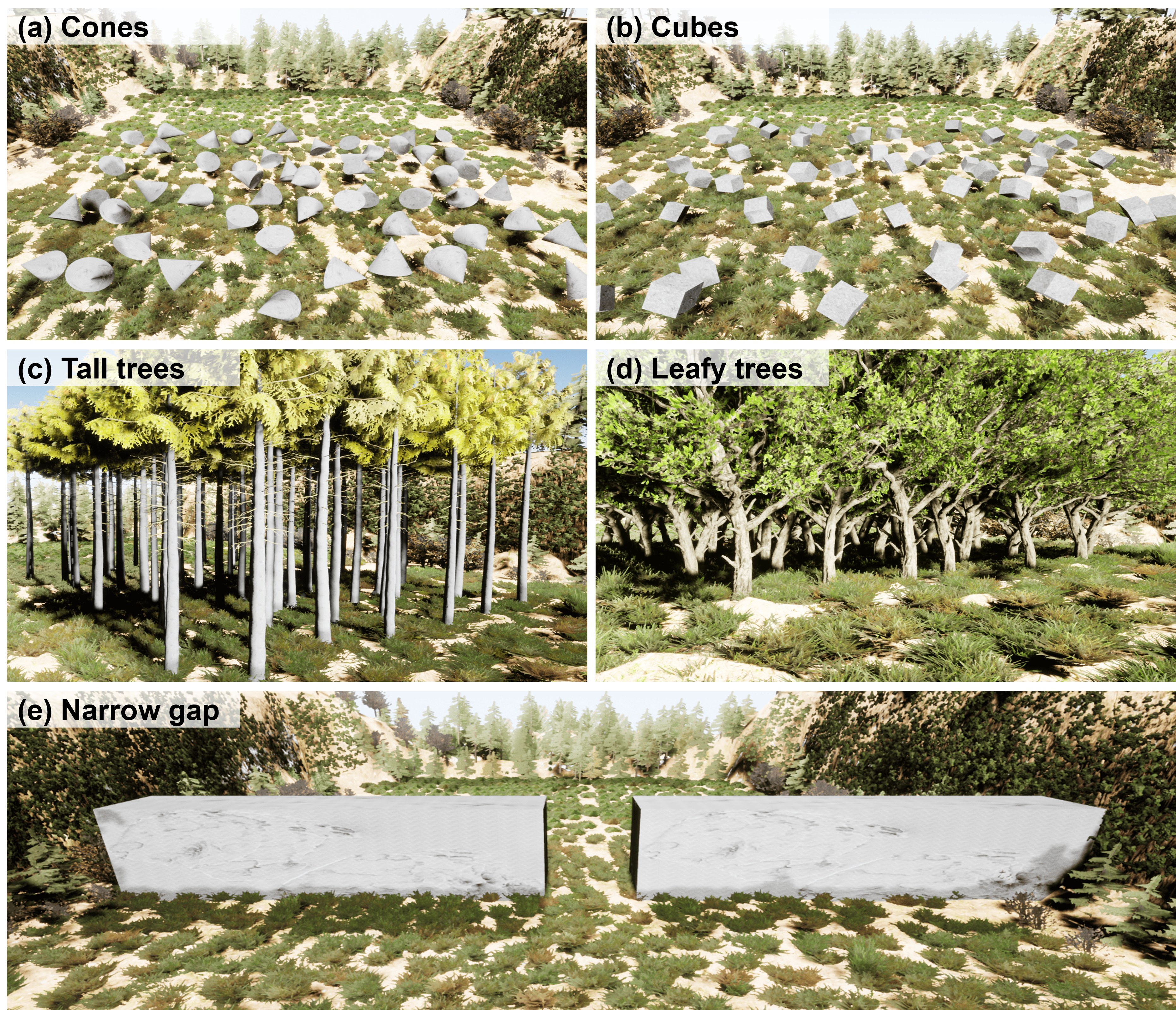}

    \caption{Various training environments in simulations. To train a model with generalization capabilities, obstacles such as cones, cubes, trees, and walls are placed in random locations with random sizes during training.}
    \label{fig:training_environments}
\end{figure}

To enhance generalization performance, a variety of training environments (e.g., trees, cones, and spheres) are generated as shown in Fig.~\ref{fig:training_environments}. The AirSim simulator~\cite{shah2018airsim} is used for map building, training, and testing the algorithm. For data acquisition, a motion primitive-based expert planner~\cite{liu2017search} is employed, which necessitates prior knowledge of the map. Point cloud data~(PCD) of the environment is first gathered to construct a global trajectory, after which local trajectories are sampled by considering an obstacle cost. Here, a global trajectory is defined as a complete path from the start to the goal constructed using map information, whereas a local trajectory is a refined short segment of the global trajectory, generated by accounting for obstacle costs. 
The overview of the data collection process can be found in Fig.~\ref{fig:training_framework}(a). 

A motion primitive-based expert generates global trajectories from random start and goal positions with a fixed altitude of 2 meters. For high-speed visual navigation, the average velocity is set to 7~m/s, with maximum velocity and acceleration capped at 8~m/s and 10~m/s\(^2\), respectively. To introduce diversity in collision-free trajectories under the same initial states, random perturbations of up to 0.3 radians are applied to the roll and yaw angles. Using this approach, we generated 1,800 global trajectories across 600 training maps. 
Based on collected global trajectories, local trajectories were sampled. On average, 60 local trajectories were obtained from each global trajectory at fixed time step intervals (i.e., 0.1s). Consequently, approximately 100,000 local trajectories, paired with corresponding state-action data, were collected in the simulation environment.

\subsubsection{Training} To further enhance generalization performance, domain randomization techniques are applied. First, in each episode, the drone's learning is initiated from a random starting position. Additionally, about 10 percent of noise is added to the drone's controller gain to introduce randomness. To enhance the robustness of the encoder during the learning process, the image random shuffling technique is used. If the drone collides with an obstacle or reaches the goal point, the episode is terminated, and the map is replaced every 5 episodes. Further details related to training hyperparameters can be found in the appendix (Section~\ref{Appendix}).

\subsection{Simulation Results}
\begin{figure*}[!ht]
    \centering
    \subfloat[Tree density: $1/80$]{%
        \includegraphics[width=0.24\textwidth]{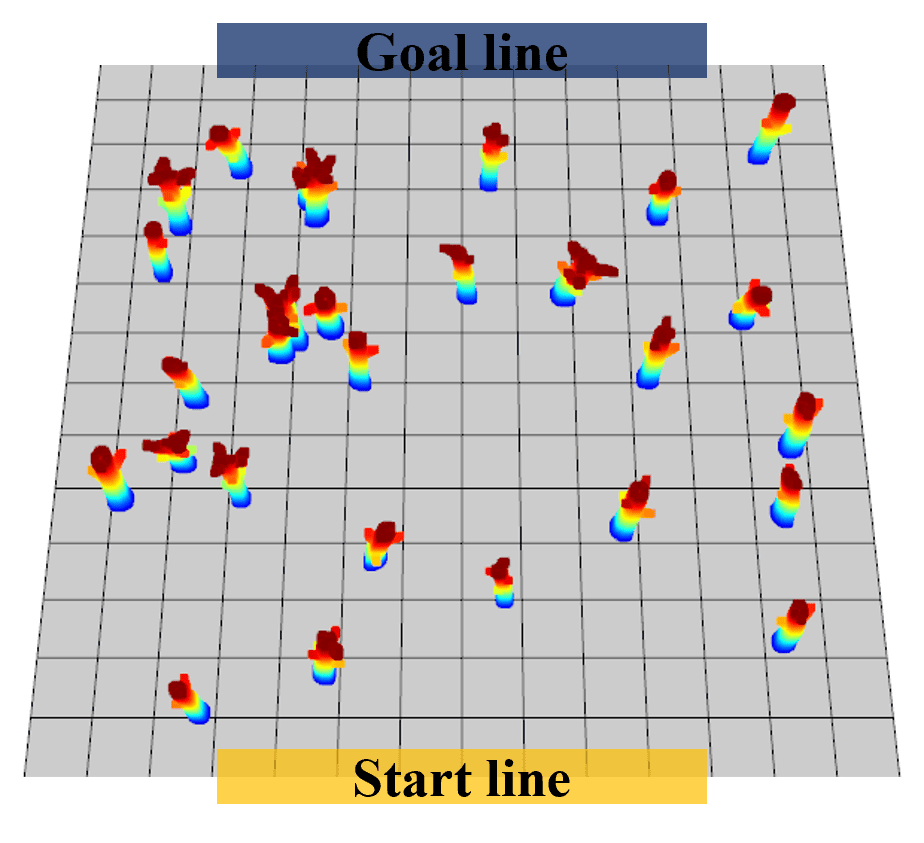}
        \label{fig:map50}
    }\hfill
    \subfloat[Tree density: $1/50$]{%
        \includegraphics[width=0.24\textwidth]{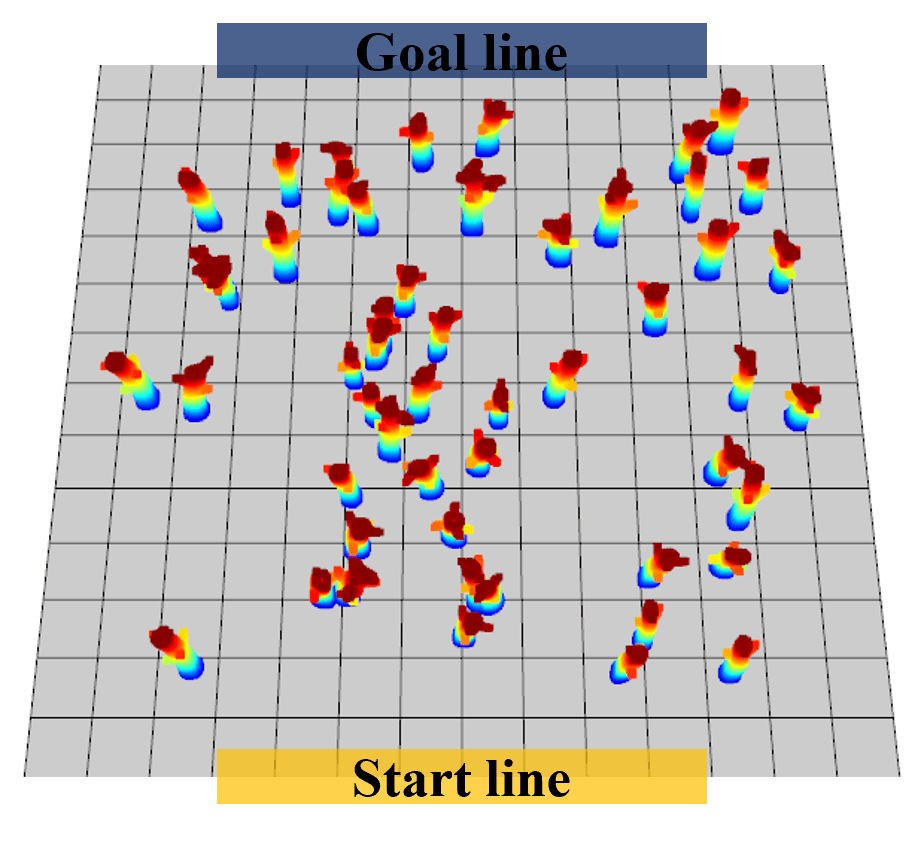}
        \label{fig:map100}
    }\hfill
    \subfloat[Tree density: $1/30$]{%
        \includegraphics[width=0.24\textwidth]{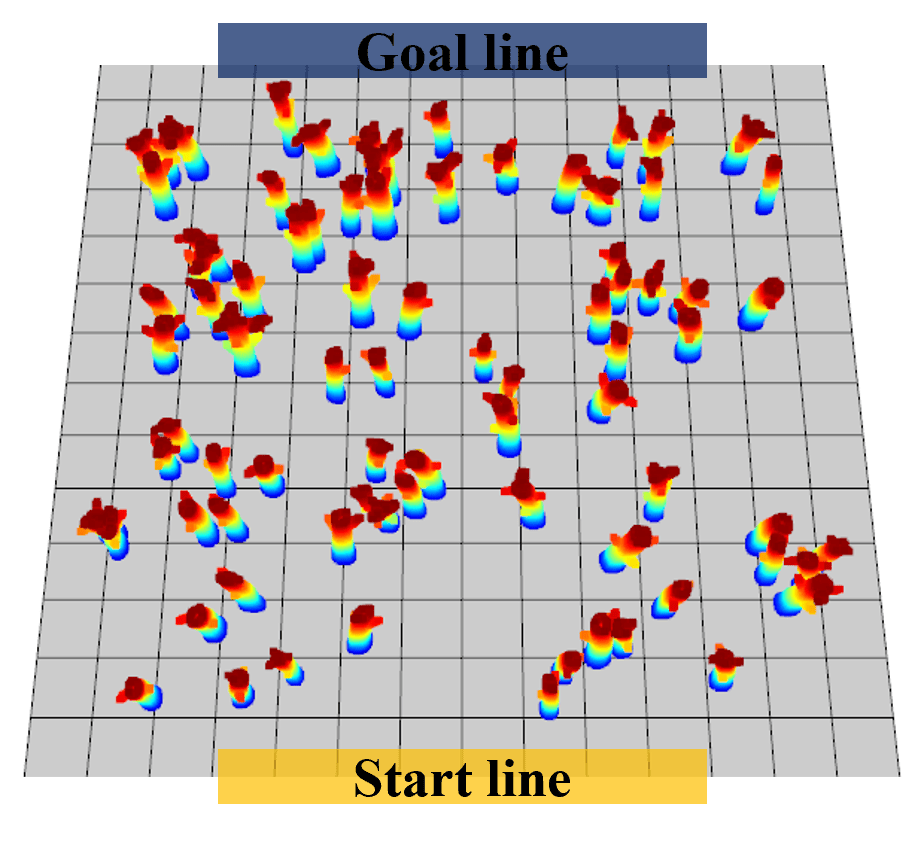}
        \label{fig:map150}
    }\hfill
    \subfloat[Tree density: $1/25$]{%
        \includegraphics[width=0.24\textwidth]{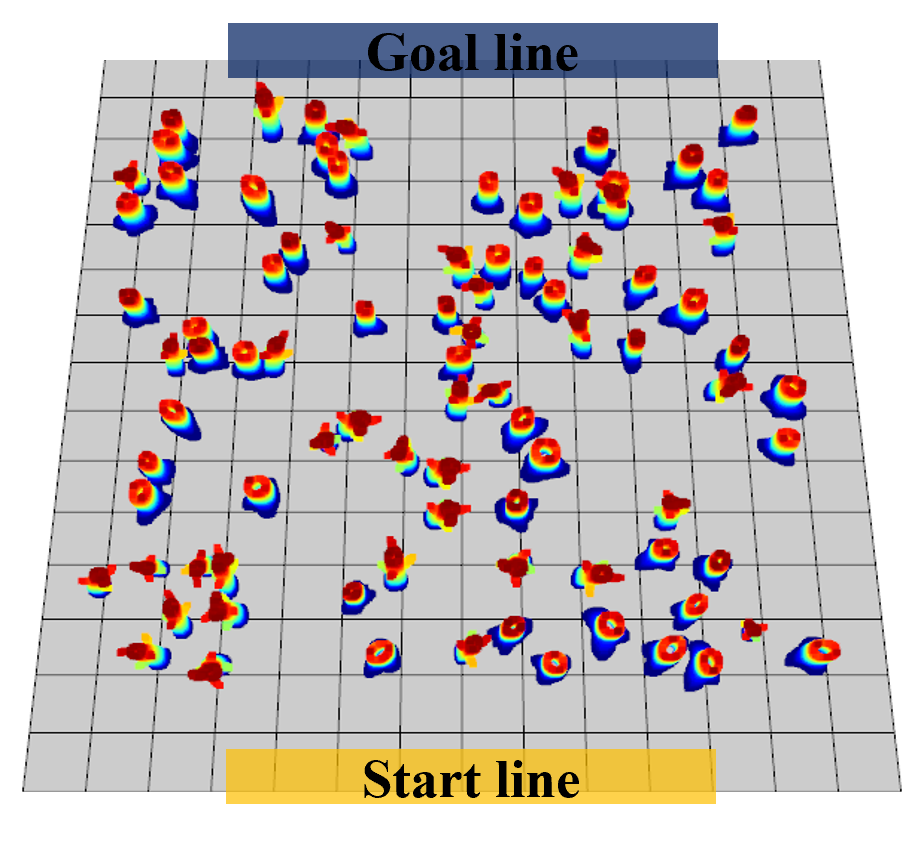}
        \label{fig:map200}
    }
    \caption{Test environments with different tree densities. Tree density represents the number of trees per unit area. The size of the grid is $5m \times 5m$.}
    \label{fig:test_environments}
\end{figure*}

\subsubsection{Comparison Methods}
To quantitatively evaluate the proposed model, we compare \textsc{RAPID} with three representative learning-based planners: a BC-based planner, an IRL-based planner, and a DAgger-based planner as well as a conventional map-based planner. Specifically, we use \textsc{BC}, \textsc{LS-IQ}~\cite{al2023ls} for IRL, \textsc{Agile}~\cite{loquercio2021learning} as the DAgger-based planner, and \textsc{Ego}~\cite{zhou2021ego} as the map-based planner. The \textsc{BC} model uses a pre-trained MobileNetV3~\cite{howard2019searching} with 1D-convolutional layers and has the same network structure as that of the \textsc{Agile}. The \textsc{LS-IQ} model shares the same network structure and hyperparameters as \textsc{RAPID}, except for the absorbing state reward update rule (detailed in Section~\ref{Policy Learning With Implicit Reward}). In particular, \textsc{LS-IQ} applies the bootstrapping method to update the Q function for both the non-terminal and the absorbing states, with the maximum and minimum absorbing reward values \(r_{A}\) set to $+2$ and $-2$, respectively. Conversely, \textsc{RAPID} combines bootstrapping for non-terminal states with an analytical update for absorbing states. Furthermore, \textsc{RAPID} uses asymmetric absorbing reward values by setting the maximum reward to $0$ and the minimum reward to $-2$, which helps to train the reward function more effectively for high-speed visual navigation tasks.

\textsc{Agile} learns collision-free trajectories using DAgger with relaxed winner-takes-all (R-WTA) loss, addressing the multi-modality issue in conventional BC methods. The original \textsc{Agile} framework employs model predictive control (MPC)~\cite{Falanga2018} to track its generated trajectories. However, in order to get a fair comparison of the waypoint generation performance under the same state inputs, we replaced the MPC with a geometric controller in this study. Although MPC can account for dynamics and enforce feasibility, the geometric controller cannot explicitly impose such constraints. To compensate for this limitation, we incorporate velocity and acceleration constraints during the trajectory generation process.

\textsc{Ego} is a representative classical map-based planner that leverages occupancy maps to avoid collisions and trajectory optimization. It is known for its robust performance in low-speed environments; however, its performance degrades at higher speeds due to modular errors and system latency. To evaluate this behavior, we used two versions of EGO-planner: a low-speed variant (\textsc{Ego-low} with a maximum velocity of 4 m/s) and a high-speed variant (\textsc{Ego-high} with a maximum velocity of 7 m/s).

\subsubsection{Validation on Test Environments}
\begin{table*}[!t]
\centering
\caption{Evaluation on different tree densities (10 trials)}
\label{table:experiment results}
\resizebox{\textwidth}{!}{%
\begin{tabular}{c|cccc|cccc}
\hline
\multirow{3}{*}{\textbf{Algorithms}} 
& \multicolumn{4}{c|}{\textbf{Avg. Mission Progress [\%] [Success/Total]}}
& \multicolumn{4}{c}{\textbf{Avg. Speed [m/s]}} \\ \cline{2-9}
& \multicolumn{4}{c|}{Tree density [trees/m\textsuperscript{2}]} 
& \multicolumn{4}{c}{Tree density [trees/m\textsuperscript{2}]} \\ \cline{2-9}
& {1/80} & {1/50} & {1/30} & {1/25} 
& {1/80} & {1/50} & {1/30} & {1/25} \\ \hline
\textsc{Ego-low} 
& 90.62 [8/10] & \textbf{88.83} [7/10] & \textbf{85.39 [7/10]} & 40.82 [0/10]
& 3.30 & 3.23 & 3.38 & 3.34 \\ \hline
\textsc{Ego-high} 
& 75.40 [5/10] & 52.56 [0/10] & 52.76 [1/10] & 34.18 [0/10]
& 5.42 & 5.19 & 5.06 & 5.45 \\ \hline
\textsc{BC} 
& 48.36 [4/10] & 43.38 [0/10] & 37.19 [0/10] & 24.27 [0/10]
& 6.50 & 6.45 & 6.47 & 5.76 \\ \hline
\textsc{LS-IQ~\cite{al2023ls}} 
& 58.26 [5/10] & 45.85 [1/10] & 32.88 [0/10] & 31.86 [0/10]
& 7.26 & 6.80 & 6.12 & 6.13 \\ \hline
\textsc{Agile~\cite{loquercio2021learning}} 
& 82.12 [6/10] & 65.25 [5/10] & 52.20 [2/10] & 52.16 [2/10]
& 5.87 & 5.57 & 5.16 & 5.52 \\ \hline
\textsc{RAPID} (Ours) 
& \textbf{94.44 [9/10]} & 87.00 \textbf{[8/10]} & 85.04 \textbf{[7/10]} & \textbf{88.69 [6/10]} 
& \textbf{7.90} & \textbf{7.32} & \textbf{7.33} & \textbf{7.27} \\ \hline
\end{tabular}%
}
\end{table*}

The experiments are carried out under varying conditions based on tree density, obstacle sizes, and shapes. Tree density indicates the number of trees per unit area. The trees are inclined and assigned random orientations to increase the complexity of the testing environment. The dimensions of trees are randomized according to a continuous uniform random distribution with $\text{scale} \sim \mathcal{U}(2/3, 4/3)$ on a $50m \times 50m$ map. Evaluation metrics include mission progress (MP), speed, and success rate (SR), where MP measures the progress made toward a goal from the starting position. In highly cluttered scenarios where the SR alone provides limited insights, MP offers a more discriminative evaluation metric. Figure~\ref{fig:test_environments} illustrates the test environments with varying tree densities. For testing, the drone starts flying from a random position within a horizontal range of $20m$ to the left or right of the map center, on the start line with an initial state of hovering. The goal point is located $60m$ directly in front of the starting position, and each method is evaluated 10 times on every map.

Table~\ref{table:experiment results} shows simulation results under varying tree densities. The classical method, EGO-planner, demonstrated high MP and success rate in low-speed settings but exhibited significantly reduced performance at higher speeds. Among various factors, the primary cause for its lower performance is planning latency and accumulated pose errors. 

The \textsc{BC} shows the lowest performance, primarily due to overfitting and compounding errors. Since \textsc{BC} strictly relies on supervised learning from the expert’s actions, any deviation from the training distribution can quickly lead to an unrecoverable error state. This distribution shift issue severely limits \textsc{BC}’s capacity to generalize, especially when the starting position varies or the environment becomes more complex.

The \textsc{LS-IQ} method performs better than \textsc{BC} but still faces notable limitations. Although it successfully mimics expert behavior in simpler simulations, \textsc{LS-IQ} tends to prioritize high-speed flight over robust collision avoidance, which leads to suboptimal performance in denser environments. Its approach to handling reward bias through absorbing states, while effective in principle, fails to fully capture the complexities of high-speed collision scenarios, resulting in diminished robustness as the tree density increases.

\textsc{Agile} demonstrates strong performance, particularly in environments with lower tree density. However, as the density and complexity grow, it exhibits an apparent performance drop. Although \textsc{Agile} effectively avoids collisions, its tendency to make large directional adjustments can paradoxically increase the likelihood of collisions in dense maps. Moreover, the method struggles to account for estimation errors of the real controller during trajectory tracking. 

\textsc{RAPID}, in contrast, achieves the best collision avoidance performance across all tested conditions, consistently surpassing the other methods in Table~\ref{table:experiment results}. Although \textsc{RAPID} initially shares the same dataset as \textsc{BC}, it gains a critical advantage by incorporating additional samples through online interactions. This online data collection not only mitigates distribution shift but also enables \textsc{RAPID} to incorporate controller tracking errors directly into the learning process. As a result, the final policy is more robust to real-world deviations, allowing for high-speed yet reliable navigation even under complex conditions.

Overall, both \textsc{Agile} (DAgger-based) and \textsc{RAPID} (IRL-based) represent robust performance for drone navigation. However, their reliance on expert data differs significantly. Whereas \textsc{Agile} requires frequent real-time labeled data from a strong expert policy, \textsc{RAPID} operates effectively even with a limited or imperfect dataset, iteratively adjusting its own data distribution to match with that of the expert. Our experiments confirm that \textsc{RAPID} not only handles constrained data conditions better but also generalizes more effectively to various environments. Consequently, \textsc{RAPID} consistently outperforms the other methods, underscoring the efficacy of inverse reinforcement learning for vision-based navigation in complex environments. Further details of the experiment results are provided in the appendix (Section~\ref{Appendix}).

\section{Experiments} \label{Experiments}
\begin{figure*}[!t]
	\centering
	\includegraphics[width=\textwidth]{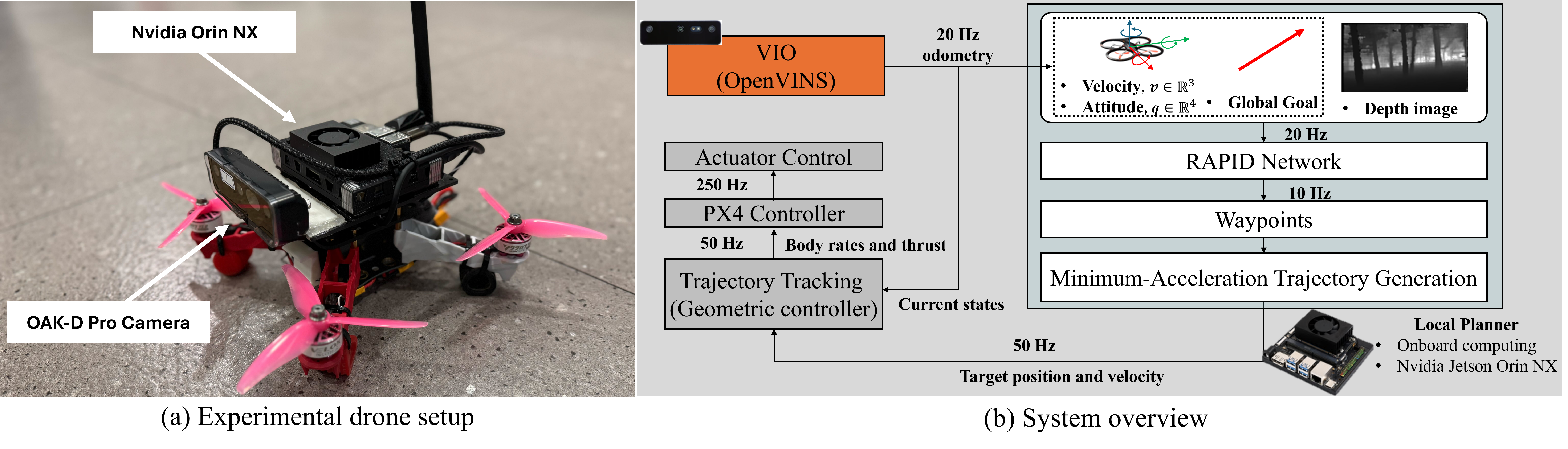}
	\caption{System overview of experimental drone.}
	\label{fig:system_integration}
\end{figure*}

\subsection{Hardware Setup}
To achieve high-speed flight, it is necessary to build a lightweight drone capable of generating powerful thrust. Therefore, we designed the hardware similar to racing drones. The drone shown in Fig.~\ref{fig:system_integration}(a) is equipped with Velox 2550 kV motors paired with Gemfan Hurricane 51466 propellers. For electronic speed controls (ESCs), we used Cyclone 45A BLHeli\_S ESCs. The overall weight of the drone is 1.1 kg and during testing, it achieved a thrust-to-weight ratio of 3.57, demonstrating its capacity for high-speed and agile maneuvers. 

For onboard computation, we employed the NVIDIA Jetson Orin NX. We deployed the neural network on the board and measured its real-time performance. Table~\ref{table:processing_latency} shows the onboard processing latency. The execution speed of the proposed model was compared with that of the \textsc{Agile}. Although the number of parameters is higher, the proposed \textsc{RAPID} model demonstrates faster execution speed due to its threefold lower FLOPS. The onboard inference test shows that the inference time of \textsc{RAPID} is more than six times faster than the \textsc{Agile}.

In this study, we used the Oak-D Pro depth camera for depth measurement and visual inertial odometry (VIO). The camera provides stereo images with an 80\textdegree{}\(\times\)55\textdegree{} field of view and stereo depth images with a 72\textdegree{}\(\times\)50\textdegree{} field of view. Both the stereo images and stereo depth images are captured at 20\,Hz. The stereo images are used for VIO state estimation, while the stereo depth images are used for the neural network input.

\begin{table}[!t]
\caption{Processing latency comparison}
\label{table:processing_latency}
\centering
\resizebox{\columnwidth}{!}{%
\begin{tabular}{|c|c|c|c|}
\hline
\textbf{Algorithm} & \textbf{Parameters} & \textbf{FLOPS\textsuperscript{*}} & \textbf{\begin{tabular}[c]{@{}c@{}}GPU inference\\ (Orin NX)\end{tabular}} \\ \hline
\textsc{Agile}     & 2.68 M             & 97 M                               & 63.89 ms                                                                  \\ \hline
\textsc{RAPID}     & 6.34 M             & 26 M                               & 10.24 ms                                                                  \\ \hline
\end{tabular}%
}
\caption*{\footnotesize \textsuperscript{*}FLOPS: Floating point operations per second, a measure of computational performance.}
\end{table}

\subsection{System Overview}
This section explains the modules of our proposed system. The proposed system mainly consists of three sub-modules: VIO, local planner, and controller. Figure~\ref{fig:system_integration}(b) shows an overview of the proposed system, including the integration of the VIO, local planner, and the controller module.

For stable high-speed flight, the VIO must be not only accurate but also robust. For our research, we use OpenVINS~\cite{geneva2020openvins}, which has been shown to perform reliably in fast flight scenarios~\cite{geneva2019}. OpenVINS takes image state information along with IMU measurements. The depth camera operates at 20 Hz, while the IMU measurements are collected at 200 Hz. This raw odometry information is integrated with the PX4 autopilot and the local odometry information is published at 20 Hz.

In the local planner module, the proposed \textsc{RAPID} method takes the depth image \textit{\textbf{I}}, velocity \textit{\textbf{v}}, attitude \textit{\textbf{q}}, and goal direction vector \textit{\textbf{g}} and then generates collision-free waypoints at 10 Hz. The average running time of \textsc{RAPID} is extremely fast (around 10 ms). The generated waypoints are then converted to a continuous trajectory using a minimum acceleration-based trajectory generation.

Given this continuous trajectory, we periodically sample it to obtain the target position and velocity commands via differentiation at each time step. In our system configuration, these commands are generated at 50~Hz. Although a higher command frequency is possible, we synchronize with the AirSim simulator to minimize the sim-to-real gap. The resulting target commands are then passed to a geometric controller for trajectory tracking. The geometric controller computes the body rates and thrust commands necessary to follow the target position and velocity. Finally, these commands are sent to the PX4 controller, which controls the drone actuators at 250~Hz.

\subsection{Experiment Results}
To validate the trained model in real-world environments, experiments were carried out in environments with two distinct characteristics: natural environments and urban environments. The evaluation focused on two main aspects: the ability to perform high-speed flight with collision avoidance and generalization performance across multiple environments. Further details regarding the experiments can be found in the supplementary video material.

\subsubsection{Natural Environments}
The experiments were carried out in natural environments divided into two scenarios: long forest and short forest. In the long forest scenario, the trees were spaced 5 meters apart and the goal point was set 60 meters away. The flight started from a hovering state, and the drone flew toward the goal, encountering obstacles along the way. During the flight, the \textsc{RAPID} approach showed obstacle avoidance movements while flying towards the goal, reaching a maximum speed of 7.5 m/s. 

\begin{figure*}
    \centering
    \includegraphics[width=\textwidth]{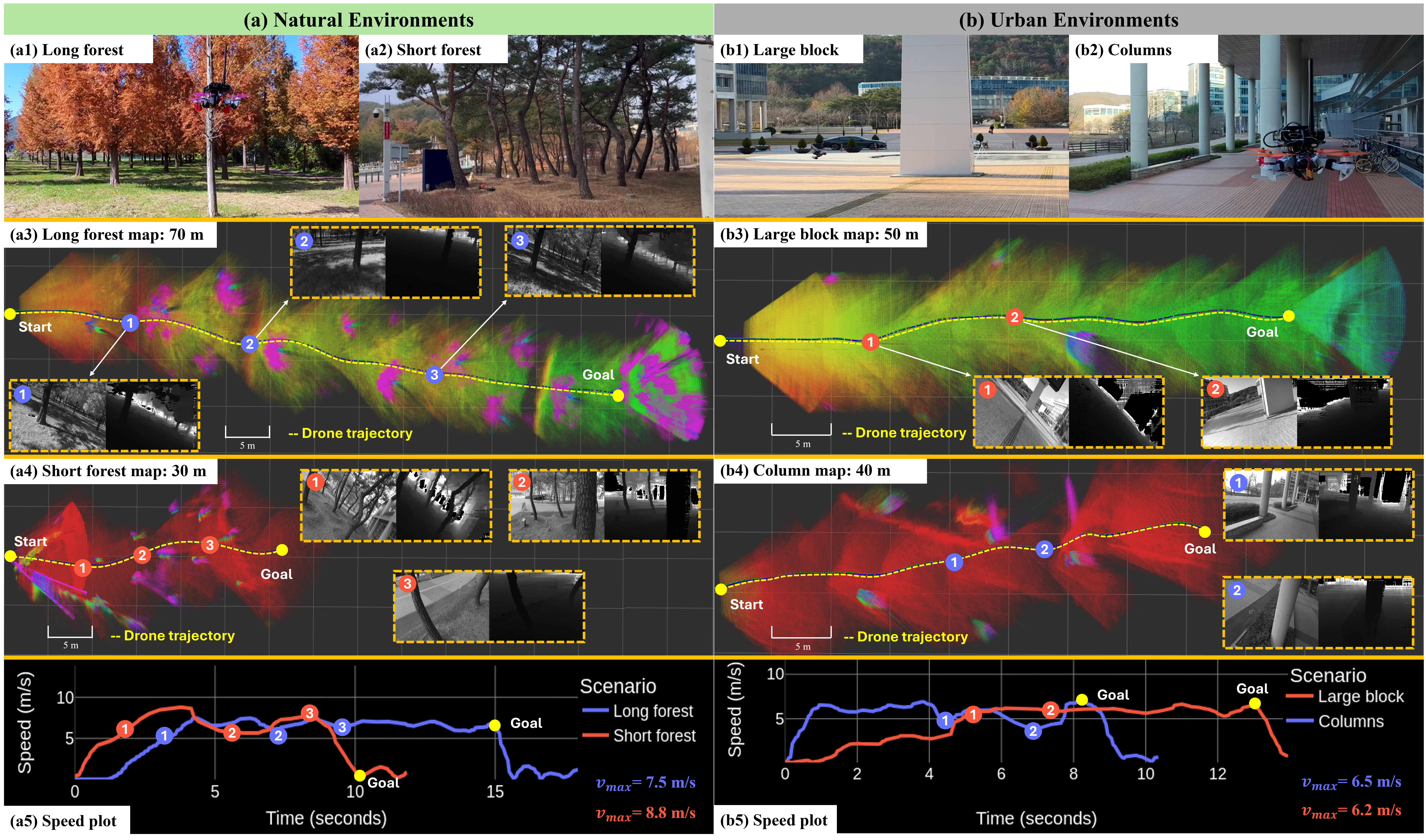}
    \caption{Flight experiment results in (a) natural and (b) urban environments. (a1–a2) and (b1–b2) show the experimental setups for each scenario. (a3–a4) and (b3–b4) present trajectory and map visualizations along with onboard grayscale/depth images. (a5) and (b5) show the corresponding speed plots.}
    \label{fig:combined environments}
\end{figure*}

We further extended the experiments to a much denser environment: the short forest. In the short forest scenario, curved trees were densely arranged within 2 meters, making the environment more complex. The goal point was set 30 meters away. In this environment, we aimed to push the drone to higher speeds to assess whether visual navigation would still be feasible under denser obstacle conditions. In the waypoint generation step, the speed at which the drone follows the planned trajectory is determined. In the long forest scenario, waypoints were generated such that the drone could follow them within 1~second. For the short forest scenario, however, this duration was reduced to 0.9~seconds to test the drone's ability to navigate through denser environments at higher speeds. Despite the increased difficulty, the drone successfully reached the goal without collisions, achieving a maximum speed of 8.8 m/s. Figure~\ref{fig:combined environments}(a) illustrates the results of the experiments conducted in the natural environment, providing an overview of each scenario. 

A noteworthy phenomenon was observed during the real-world experiments. Although the expert dataset used for training was collected with a constant velocity of 7~m/s, the IRL training enabled the policy to exhibit acceleration and deceleration behaviors that were not present in the dataset. In some cases, the drone even reduced its speed significantly before executing an avoidance maneuver near obstacles. This suggests that the IRL-based method goes beyond simply mimicking the expert's behavior, effectively capturing the intention of collision avoidance and integrating it into the policy.

\subsubsection{Urban Environments}
The urban environments are divided into two scenarios: a large block and columns. Figure~\ref{fig:combined environments}(b1) shows the large block environment, where obstacles are geometrically simple yet significantly large. In these urban settings, we deliberately operate the drone at slightly lower speeds due to higher safety risks. Unlike forest environments, where crashes typically cause minimal damage thanks to softer terrain, an impact on the hard brick surfaces found in urban environments can severely damage the drone. Therefore, the drone's speed is intentionally reduced to an average of 6~m/s to mitigate the risk of major damage. This environment also requires the drone to generate avoidance trajectories at an early stage. As shown in Fig.~\ref{fig:combined environments}(b3), the drone successfully generated an avoidance path from the very beginning (point~1) and reached the destination achieving a maximum speed of 6.2~m/s.

Figure~\ref{fig:combined environments}(b2) describes the experimental setup with column-shaped obstacles, consisting of six large columns. Figure~\ref{fig:combined environments}(b4) shows the flight trajectory in the column environment. Similar to previous experiments, the drone decelerated while avoiding obstacles and accelerated again once it passed the obstacles, reaching the destination successfully. In the column environment, the drone reached a maximum speed of 6.5 m/s. 

From the experiments conducted in both natural and urban environments, although the model was trained in a simulation environment, it could achieve good performance in real-world scenarios with minimal performance degradation compared with the simulation environment. Experiment results in natural environments indicate that the sim-to-real gap has been partially mitigated when testing in tree environments similar to the simulation setting (Fig.~\ref{fig:training_environments}(c) and Fig.~\ref{fig:training_environments}(d)). Furthermore, experiments in urban environments demonstrated the model's ability to generalize to new obstacle shapes, highlighting its adaptability to diverse real-world scenarios.

\section{Limitations} \label{Limitation}
Although the proposed method has shown good performance in simulations and real-world environments, several challenges still remain. 

\subsection{Lack of Temporal Awareness}
The \textsc{RAPID} algorithm generates collision-avoidance trajectories based on a single image and the current drone state. While this approach provides the advantage of generating instantaneous avoidance paths, it lacks the history of previously avoided obstacles. As a result, the algorithm is prone to falling into local minima, particularly when encountering large obstacles (e.g., walls). In such scenarios, an incorrect initial avoidance trajectory can prevent the drone from navigating out of the environment, ultimately leading to collisions. To address this issue, one potential solution is to incorporate multiple sequential images as an input to capture temporal information for trajectory generation and leverage long-short term memory networks~\cite{yu2024mavrl, bhattacharya2024vision, nguyen2024uncertainty}. However, both methods may introduce inference delays compared with the simple convolutional layers, requiring careful consideration to balance computational efficiency and performance.

\subsection{Generating Infeasible Trajectories During Exploration} 
A key challenge in generating collision-free trajectories at high-speed is ensuring that the drone only produces physically feasible trajectories. Achieving this typically demands considerable exploration, which inevitably generates many infeasible trajectories during the early stages of training. Excessive exploration can prevent the Q-function from converging, destabilizing the learning process. Conversely, reducing exploration to avoid infeasible trajectories often leads to a suboptimal policy, resulting in lower overall performance.

To address this issue, trajectory generation methods that incorporate velocity and acceleration constraints~\cite{me2011minimum}, as well as feasible trajectory tracking controllers like MPC, could be considered. In the trajectory generation process, rather than using a closed-form method, there is a constrained optimization approach that can yield a smoother path by considering velocity and acceleration constraints, even if it does not pass exactly through each waypoint. The MPC generates safe and feasible trajectories by solving an optimization problem within given constraints. However, both methods involve computational complexity and reliance on solving optimization problems in real-time, making them less suitable for RL training. These limitations underscore the need for a more fundamental, learning-based solution that can effectively address these challenges within the RL framework.

To alternatively address this problem, the agent's search space was constrained to closely resemble that of the expert, as discussed in Section~\ref{States and Actions}. This was achieved by designing the action space using cylindrical coordinates, which were specifically chosen to inherently restrict the exploration space and better align it with the expert's behavior. However, while this approach mitigates some issues, it does not provide a fundamental solution. Other prior studies have explored the application of constraint-RL methods as an alternative approach~\cite{kim2024not}. By penalizing infeasible actions during training, such approaches are expected to effectively guide the agent toward robust and feasible solutions. Another potential solution involves using a warm-up phase with an initial policy~\cite{xing2024bootstrapping}. Employing a warm-up policy significantly reduces the action space, allowing the generation of feasible waypoints. However, this method requires careful balancing between the actor and critic, making it challenging to apply in IRL settings. Significant efforts would be required to adapt these techniques to practical use in IRL.

\subsection{Episodic Incompleteness in Expert Dataset}
During the data acquisition process, the SE(3) planner was used to generate smooth trajectories that the drone could smoothly follow in a given map. To sample obstacle avoidance trajectories, additional obstacle-related costs were incorporated into the existing trajectory, and random sampling was performed to acquire the dataset. However, the resulting dataset is not ideal for IRL. 

For IRL, the state-action pairs need to form a complete trajectory from the initial state to the terminal state. In this process, random sampling for obstacle avoidance often leads to a break in the trajectory, resulting in incomplete episodes. Consequently, when the trained model encounters states far from the expert trajectory, it fails to find a solution and struggles to achieve further performance improvements. Moreover, in such states, the multi-modality issue arises, where multiple avoidance paths can exist for a single state, further complicating the problem~\cite{tordesillas2023deep}. To address this issue, states following such out-of-distribution events need to be collected to form complete episodes. However, this aspect remains unresolved, posing a significant limitation in the current data acquisition methodology.

\subsection{Imperfect Mitigation of Sim-to-Real Gap}
To address the sim-to-real gap, a stereo vision algorithm (i.e., SGM) was employed to reflect the characteristics of a real stereo depth camera (Section \ref{States and Actions}). However, during real-world experiments, it was observed that discrepancies in hardware specifications between the trained drone and the actual drone led to differences in controller tracking performance, thereby introducing an additional sim-to-real gap.

To partially mitigate this issue, domain randomization of the geometric controller's gain values was applied during training. While this approach improved robustness to some extent, it failed to completely resolve the sim-to-real gap. A more fundamental solution would involve training across multiple drones with varying sizes and dynamics in parallel. Such an approach can enhance the robustness of trajectory planning and improve the transferability of the learned policy~\cite{nahrendra2023dreamwaq}.

\section{Conclusions and Future Work} 
\label{Conclusions}
This study proposed \textsc{RAPID}, a robust and agile planner for drones utilizing inverse reinforcement learning to achieve high-speed visual navigation in cluttered environments. Unlike conventional methods, \textsc{RAPID} integrates visual inputs (i.e., perception) and planning to generate collision-free waypoints in real time, demonstrating superior performance in both simulation and real-world scenarios. By leveraging inverse soft Q-learning with proper absorbing state treatment and auxiliary loss functions for autoencoder, \textsc{RAPID} effectively addressed challenges in sample efficiency and high-dimensional visual input processing. The trained policy showed good generalization capabilities, achieving average and maximum speeds of 7 m/s and 8.8 m/s, in natural and urban environments respectively without additional real-world training or tuning.

While \textsc{RAPID} performed robustly, several limitations remain. Lack of temporal awareness makes large‑obstacle avoidance difficult, indicating the need for sequential or memory‑based perception. The challenge of generating feasible trajectories underscores the importance of refining sampling strategies or using constraint RL. Finally, the sim‑to‑real gap, though partly mitigated via domain randomization and stereo vision, still calls for hardware‑diverse training.

Future work will focus on addressing these limitations by exploring memory-based architectures, incorporating constraint-based reinforcement learning methods, and enhancing data acquisition strategies to enable more robust, scalable, and efficient learning for high-speed drone navigation in real-world scenarios.

\section*{Acknowledgments}
This research was supported by National Research Foundation of Korea (NRF) grant funded by the Korean government (MSIT) (2023R1A2C2003130), Basic Science Research Program through the National Research Foundation of Korea (NRF) funded by the Ministry of Education (2020R1A6A1A03040570), Unmanned Vehicles Core Technology Research and Development Program through the NRF, and Unmanned Vehicle Advanced Research Center (UVARC) funded by the Ministry of Science and ICT, the Republic of Korea (2020M3C1C1A01082375).

\section{Appendix} \label{Appendix}
\subsection{Full Q-Function Objective With Terminal State Treatment} \label{Full objective of Q with terminal state treatment}
The full objective of Q-function including terminal state treatment is expressed as: 
\begin{equation} \label{eq:full objective with terminal state}
\begin{aligned}
J(Q, \pi)& = \mathbb{E}_{d_{\pi_E}}[r(s,a) - \alpha(r(s,a))^2] \cdot (1 - \nu) \\
		 & - (1 - \alpha) \mathbb{E}_{d_{\pi_L}} [(r(s,a))^2] \cdot (1 - \nu) \\
		 & - \mathbb{E}_{d_{\pi_L} \cup d_{\pi_E}}[(V^\pi(s) - \gamma \mathbb{E}_{s' \sim P(\cdot|s,a)} V^\pi(s'))] \cdot (1 - \nu) \\
		 & + \mathbb{E}_{d_{\pi_E}} [(Q(s,a) - \frac{r_{max}}{1 - \gamma}) - \alpha (Q(s,a) - \frac{r_{max}}{1 - \gamma})^2] \cdot \nu \\
		 & - (1 - \alpha) \mathbb{E}_{d_{\pi_L}} [(Q(s,a) - \frac{r_{min}}{1 - \gamma})^2] \cdot \nu \\
		 & - \mathbb{E}_{d_{\pi_L} \cup d_{\pi_E}}[V^\pi(s) - \frac{r_A}{1 - \gamma}] \cdot \nu,
\end{aligned}
\end{equation}
where $r(s,a) = Q(s,a) - \gamma \mathbb{E}_{s' \sim P(\cdot|s,a)} V^\pi(s')$ and $\nu$ is an indicator variable, such that $\nu = 1$ if $s'$ is a terminal state and $\nu = 0$ otherwise. In this formulation, $r_{\text{max}} = 0$ and $r_{\text{min}} = -2$. In the last term of Eq.~(\ref{eq:full objective with termial state}), $r_A$ refers to $r_{\text{max}}$ if states are sampled from the expert distribution, and $r_A$ is $r_{\text{min}}$ otherwise. Our formulation builds upon the Q-function objective proposed in~\cite{al2023ls}, incorporating explicit terminal state handling.

\subsection{Actor and Critic Networks} 
This study employs SAC with a single critic network to address reward ambiguity, instead of the standard double Q-learning approach. Following the D2RL architecture~\cite{sinha2020d2rl}, both the critic and actor networks use a four-layer MLP with \texttt{LeakyReLU} activation and a hidden dimension of 1,024.

\subsection{Encoder and Decoder Networks}
We utilize 3$\times$3 kernels with 32 channels across all convolutional layers, maintaining a stride of 1 for all layers except the first, which has a stride of 2. The output of the convolutional network is then passed through a single fully-connected layer. In the original study~\cite{yarats2021improving}, \texttt{LayerNorm}~\cite{Ba2016LayerNormalization} was employed for input normalization, which proved effective in their experimental setup. However, in this study, we observed that \texttt{LayerNorm} negatively impacted learning performance. This difference arises because the original work used three consecutive frames as input, allowing \texttt{LayerNorm} to capture temporal correlations across frames, which enhanced performance. In contrast, our method uses a single image as an input, and applying \texttt{LayerNorm} in this context, combined with other network inputs, led to degraded performance. Lastly, we apply the \texttt{tanh} activation function to the 128-dimensional output of the fully-connected layer.

Both the actor and critic networks employ separate convolutional encoders. Although the convolutional layers are weight-shared between them, only the critic optimizer is allowed to update these shared weights (as shown in Fig.~\ref{fig:training_framework} and Fig.~\ref{fig:auxiliary loss}), as we block the actor’s gradient flow from affecting the convolutional layers.

The decoder architecture includes a fully-connected layer followed by four transposed convolutional layers. Each transposed convolution uses 3$\times$3 kernels with 32 channels and a stride of 1, except for the final layer, which has a stride of 2. \texttt{LeakyReLU} activations are applied after all layers except for the last transposed convolutional layer, which outputs the pixel representation. The critic’s encoder and the aforementioned decoder are combined into an autoencoder. Notably, because the convolutional layers are shared between the encoders of the actor and the critic, the reconstruction signal from the autoencoder also influences the convolutional layers in the actor’s encoder.

\subsection{Weights Initialization}
Fully-connected layers are initialized with orthogonal weights~\cite{saxe2013exact} and zero bias, while convolutional layers use delta-orthogonal initialization~\cite{xiao2018dynamical}.

\subsection{Hyperparameters}
Hyperparameters for the training are summarized in Table~\ref{tab:hyperparameters}.

\begin{table}[H]
    \centering
    \caption{Hyperparameters used in the experiments.}
    \label{tab:hyperparameters}
    \begin{tabular}{ll}
        \hline
        \textbf{Parameter name}                 & \textbf{Value}         \\ \hline
        Number of convolutional layers          & 4                      \\
        Number of nodes for MLP                 & 1,024                   \\
        Replay buffer capacity                  & 600,000                \\
        Batch size                              & 128                    \\
        Encoder embedding feature dim.          & 128                    \\
        Random initial action                   & 1,000                  \\        
        Discount $\gamma$                       & 0.99                   \\
        Optimizer                               & Adam                   \\
        Temperature Adam’s $\beta_1$            & 0.5                    \\
        Init temperature                        & 0.1                    \\        
        Critic Q-function soft-update rate $\rho_Q$ & 0.005            \\
        Critic encoder soft-update rate $\rho_{\text{enc}}$ & 0.01    \\
        Critic target update frequency          & 1                      \\
        Actor update frequency                  & 1                      \\
        Actor log stddev bounds                 & $[-10, 2]$             \\
        Critic learning rate                    & $3 \times 10^{-4}$     \\
        Autoencoder learning rate               & $3 \times 10^{-4}$     \\
        Actor learning rate                     & $3 \times 10^{-5}$     \\        
        Temperature learning rate               & $3 \times 10^{-5}$     \\ 
		Decoder latent $\lambda_{z}$            & $10^{-7}$ 			   \\
		Weight decay of decoder $\lambda{\theta}$& $10^{-7}$             \\       
        \hline
    \end{tabular}
\end{table}
\subsection{Additional Simulation Results}
\begin{figure}[h]
    \centering
    \subfloat[Ego-low]{%
        \includegraphics[width=0.115\textwidth]{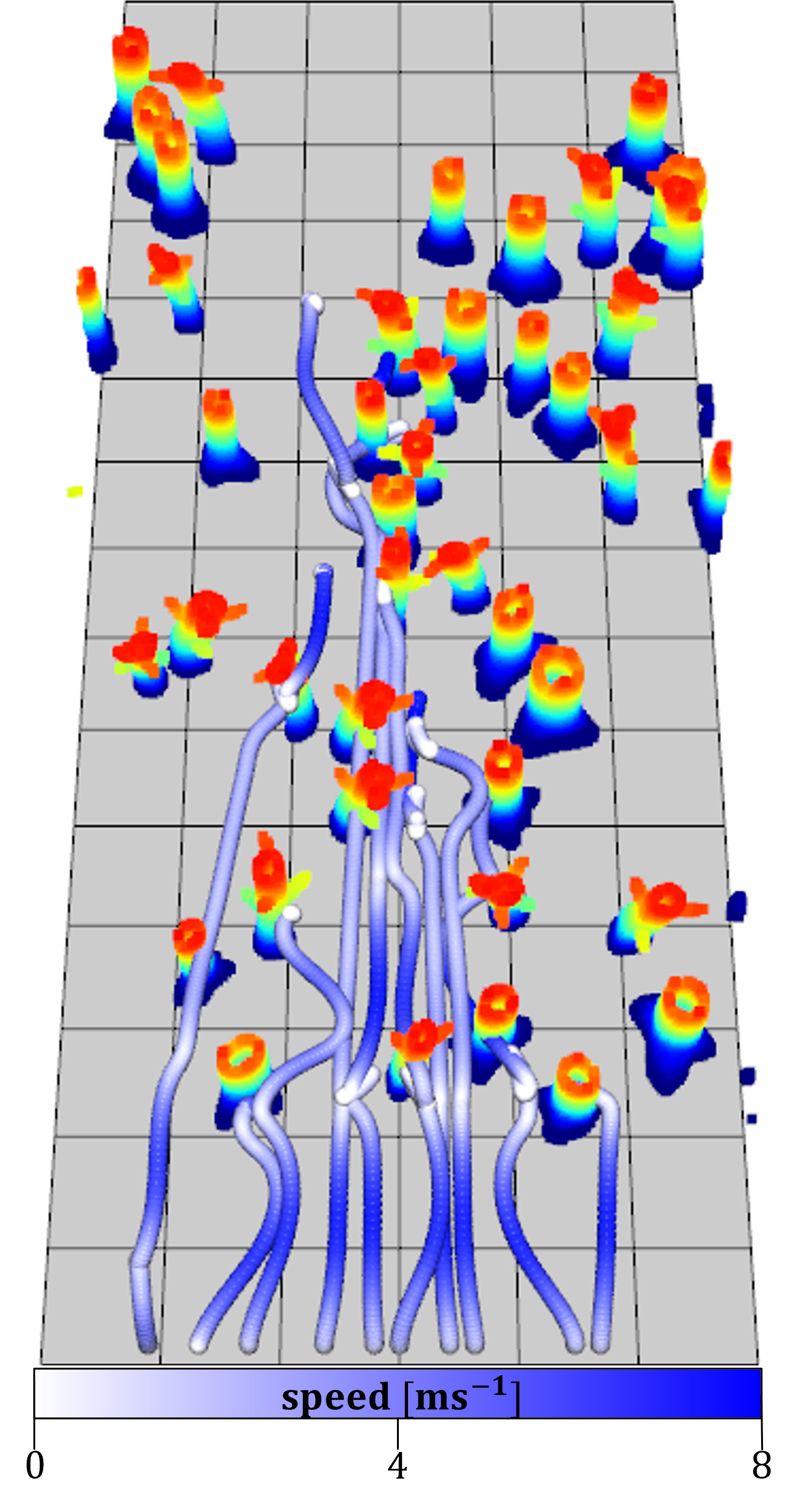}
    }
    \subfloat[Ego-high]{%
        \includegraphics[width=0.115\textwidth]{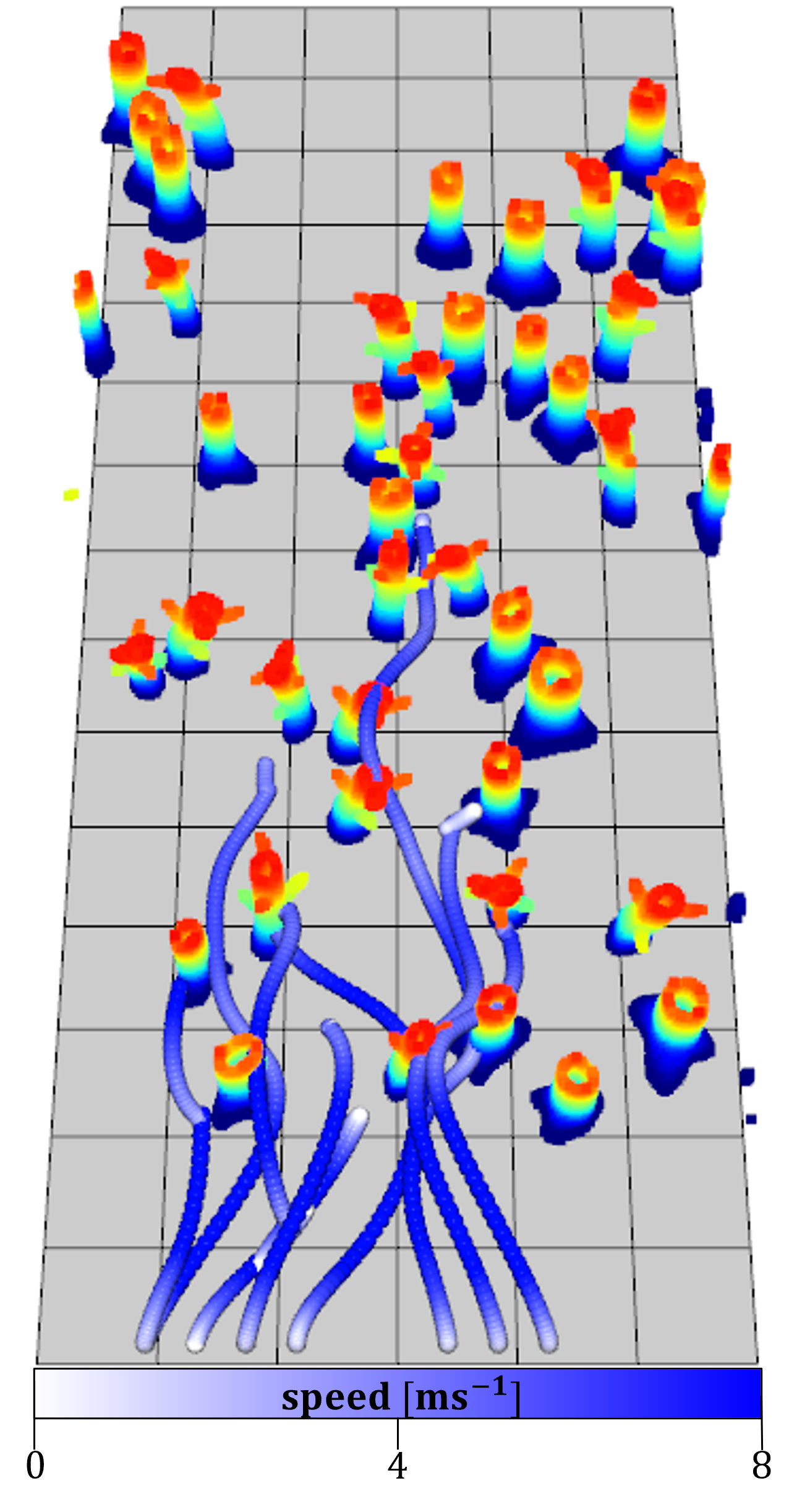}
    }
    \subfloat[Agile]{%
        \includegraphics[width=0.115\textwidth]{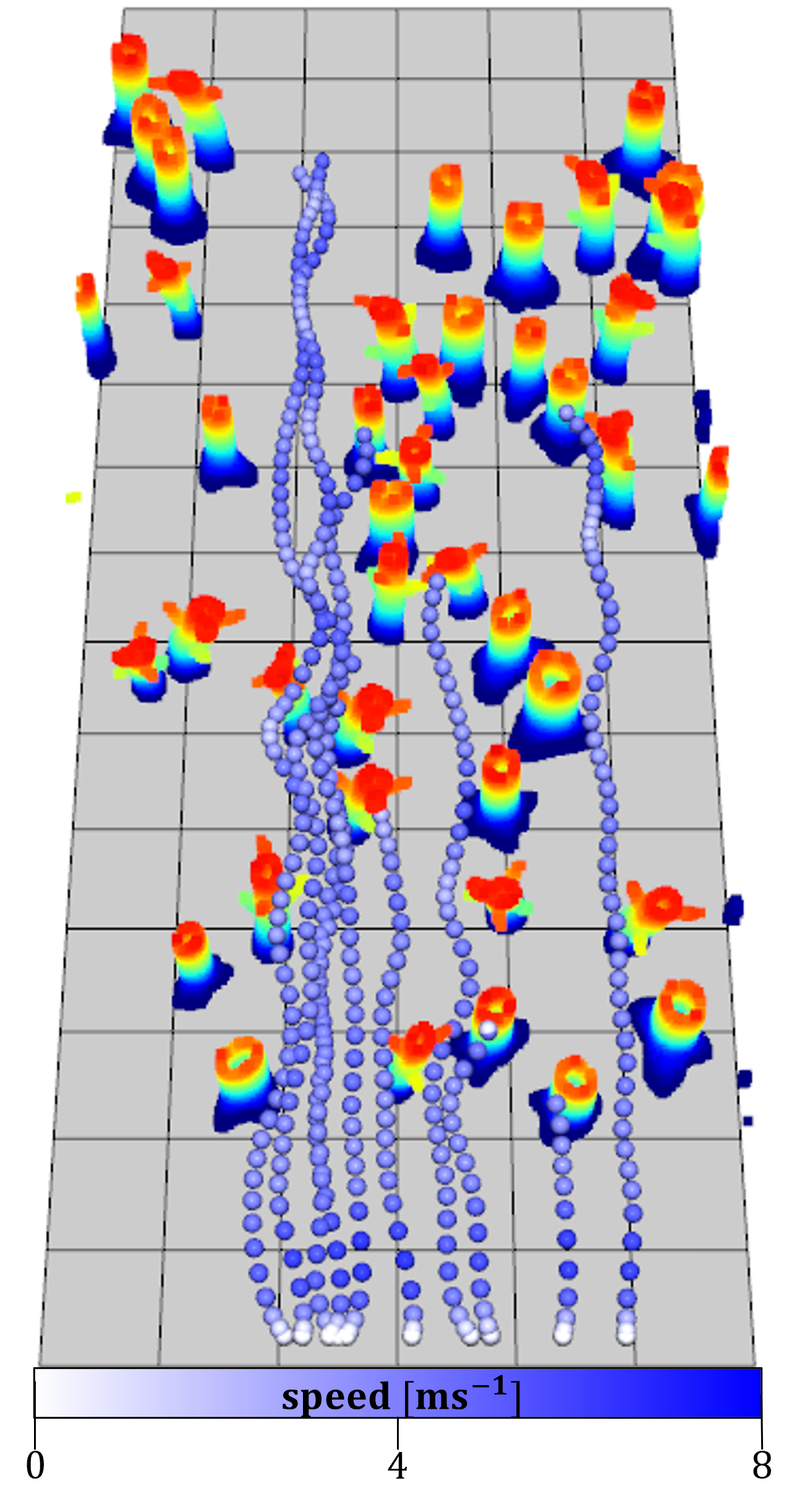}
    }
    \subfloat[RAPID]{%
        \includegraphics[width=0.115\textwidth]{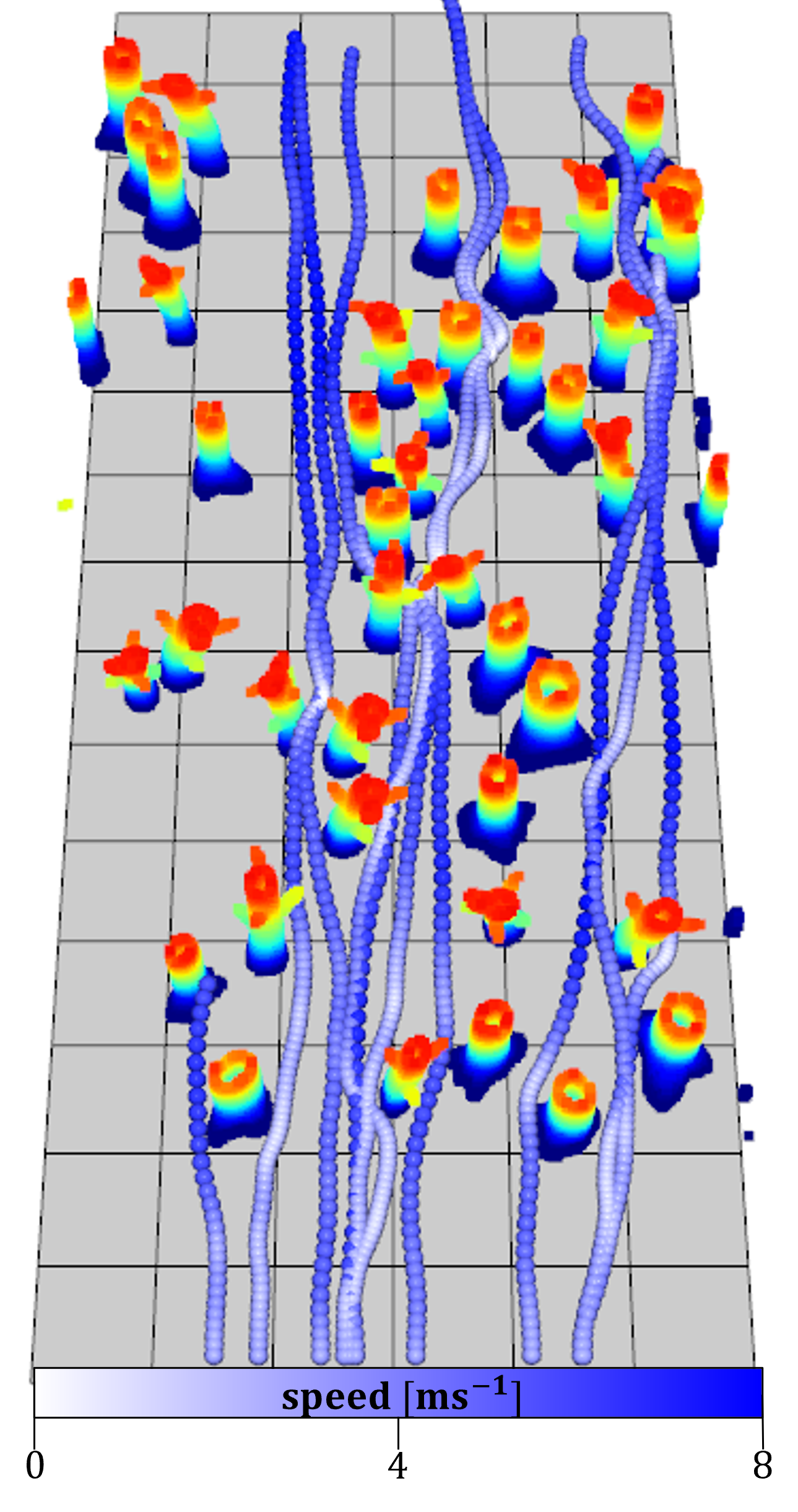}
    }
    \caption{Flight trajectories with speed in a map with a tree density of 1/25 (10 trials).}
    \label{fig:flight history}
\end{figure}

Figure~\ref{fig:flight history} visualizes flight trajectories and corresponding speeds in a map with a tree density of 1/25. Note that tree branches are omitted for clear presentation. EGO-planner often became stuck mid-flight due to delayed or failed planning. In contrast, both \textsc{Agile} and \textsc{RAPID} adapted to new scenes due to their fast inference capability. Particularly, \textsc{RAPID} adaptively flew at high speeds in open areas and reduced speeds in dense obstacle regions. Despite being trained with an expert dataset collected at high speeds, \textsc{RAPID} effectively learned diverse collision-avoidance maneuvers suitable for varying flight conditions.

\end{document}